\documentclass[journal]{IEEEtran}
\IEEEoverridecommandlockouts                              

\usepackage{times}
\usepackage{multicol}

\usepackage{graphicx}
\usepackage{algorithm}
\usepackage{algpseudocode}
\usepackage{amsmath, amssymb, mathtools}
\usepackage{cancel}
\usepackage[most]{tcolorbox}
\usepackage[table]{xcolor} 
\usepackage{xcolor}
\usepackage{tikz}
\usetikzlibrary{shapes, arrows.meta, positioning, shadows, calc, fit}
\usepackage{booktabs} 
\usepackage{multirow} 
\usepackage[normalem]{ulem}
\usepackage{flushend}
\usepackage{balance}
\usepackage{makecell}
\usepackage[cal=boondoxo]{mathalfa}
\usepackage{cuted}   
\usepackage{capt-of}  
\usepackage{subcaption}
\usepackage{graphicx}

\usepackage[most]{tcolorbox}
\tcbset{
  promptbox/.style={
    enhanced,
    breakable,
    colback=gray!5,
    colframe=black!15,
    boxrule=0.4pt,
    arc=1.2mm,
    left=1.2mm,right=1.2mm,top=1.0mm,bottom=1.0mm,
  }
}

\usepackage[colorlinks,bookmarks=true,linkcolor=blue,citecolor=blue,urlcolor=blue]{hyperref}
\usepackage[figure,figure*,table,table*]{hypcap} 

\algnewcommand{\LineComment}[1]{\Statex \hfill \textcolor{gray!90}{\footnotesize \textit{$\triangleright$ #1}}}

\begin{document}
\bstctlcite{IEEEexample:BSTcontrol}
%
\title{CoFL: Continuous Flow Fields for Language-Conditioned Navigation}
%
%

\author{Haokun~Liu, Zhaoqi~Ma, Yicheng~Chen, Masaki~Kitagawa, Wentao~Zhang, Zicen~Xiong, Jinjie~Li, Moju~Zhao
}
\maketitle
\begin{strip}
    \vspace*{-2.0cm}
    \centering
    \includegraphics[width=\textwidth]{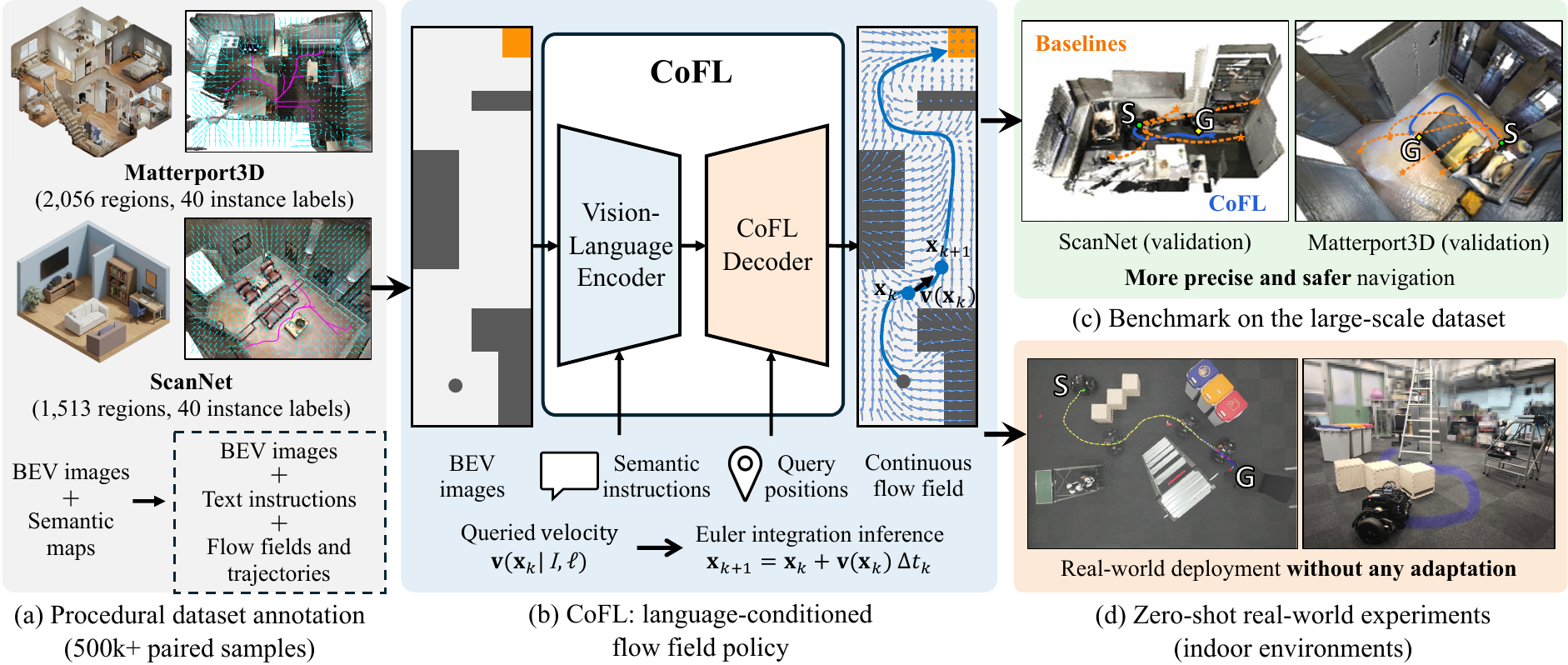}  
    \captionof{figure}{{Overview of the main contributions.} (a) An automated annotation pipeline that constructs a large-scale BEV image--instruction dataset with procedurally generated flow-field and trajectory references (500k+ samples) from indoor 3D scenes. (b) CoFL: a transformer-based language-conditioned policy that predicts continuous flow fields, enabling smooth trajectory rollout with a controllable inference budget. (c) Benchmark results on unseen environments, demonstrating more precise and safer navigation. (d) Zero-shot transfer to real-world robot navigation using a model trained only on the proposed dataset.}
    \label{fig:overview}
    \vspace{-3mm}
\end{strip}


\begin{abstract}
Existing language-conditioned navigation systems typically rely on modular pipelines or trajectory generators, but the latter use each scene--instruction annotation mainly to supervise one start-conditioned rollout.
To address these limitations, we present CoFL, an end-to-end policy that maps a bird's-eye view (BEV) observation and a language instruction to a continuous flow field for navigation. 
CoFL reformulates navigation as workspace-conditioned field learning rather than start-conditioned trajectory prediction: it learns local motion vectors at arbitrary BEV locations, turning each scene--instruction annotation into dense spatial control supervision.
Trajectories are generated from any start by numerical integration of the predicted field, enabling simple real-time rollout and closed-loop recovery.
To enable large-scale training and evaluation, we build a dataset of over 500k BEV image--instruction pairs, each procedurally annotated with a flow field and a trajectory derived from semantic maps built on Matterport3D and ScanNet. 
Evaluating on strictly unseen scenes, CoFL significantly outperforms modular Vision-Language Model (VLM)-based planners and trajectory generation policies in both navigation precision and safety, while maintaining real-time inference. 
Finally, we deploy CoFL zero-shot in real-world experiments with BEV observations across multiple layouts, maintaining feasible closed-loop control and a high success rate.
\end{abstract}


%
\IEEEpeerreviewmaketitle

\section{Introduction}
\label{sec:introduction}

Language-conditioned navigation seeks to translate high-level semantic instructions into low-level, continuous motion in complex environments.
Despite rapid progress, existing pipelines remain fragmented. 
Modular approaches decompose the problem into perception, grounding, and planning components, which improves interpretability but can be brittle to upstream errors.
End-to-end policies avoid such hand-designed interfaces by directly predicting actions or trajectories from observations and language, making them appealing for scalable vision-language control.

However, this action-prediction formulation also determines how supervision is used.
Recent end-to-end robot policies are commonly formulated as predicting an action token or an action chunk conditioned on the current observation and task instruction~\cite{Brohan-RSS-23,zitkovich2023rt,kim2024openvla,chi2025diffusion,BlackK-RSS-25,black2025pi}. 
In these formulations, each training target corresponds to the behavior demonstrated from the state at which the example was collected. 
Thus, each observation--instruction pair supervises the demonstrated rollout, rather than the broader spatial behavior induced by the same instruction in the same scene. 
This underuses the annotation for navigation: a scene and a target define useful local guidance from many feasible locations, but trajectory supervision reveals only the states visited by one rollout. 

This motivates a representation that can expose such workspace-level motion patterns as dense supervision.
Instead of treating an annotation as a single start-anchored rollout, one can define local control targets over the workspace: for any queried location, the policy should predict how the robot ought to move under the given observation and instruction. 
Such a field-based formulation turns each scene--instruction pair into dense spatial supervision, decoupling the learned guidance from one particular initial state while still allowing trajectories to be recovered by integration.

Bird’s-eye view (BEV) and BEV-aligned elevated representations offer a lightweight yet globally consistent workspace in which free space, obstacles, starts, and goals are represented in a common metric frame. 
This property makes BEV-aligned workspace not only convenient for execution, but also well suited for studying field-based control: a single scene can be queried at many spatial locations, allowing supervision and evaluation beyond one demonstrated start.

This raises a natural question: can a vision-language policy learn such dense workspace-level guidance directly from BEV observations and language instructions, while retaining the simplicity of end-to-end inference?
We propose CoFL, a transformer-based policy that reformulates language-conditioned navigation in the BEV-aligned workspace as flow field estimation. 
CoFL learns a navigation field $\mathbf{v}(\mathbf{x}\mid I,\ell)$ that maps an observation $I$, an instruction $\ell$, and a queried workspace location $\mathbf{x}$ to a local motion vector. 
Unlike trajectory policies that predict a single future rollout from a specified start, CoFL learns a spatially queryable control landscape over the workspace domain, including states not visited by the reference trajectory. 
At inference time, trajectories are obtained by numerical integration of the predicted field from the current robot state, enabling real-time closed-loop execution and recovery from off-trajectory states.

We summarize our contributions as follows (Fig.~\ref{fig:overview}):
\begin{enumerate}
    \item We introduce CoFL, a field-based policy that reformulates language-conditioned navigation from start-conditioned trajectory prediction into dense workspace-level control learning. The learned field can be queried at arbitrary locations and integrated into continuous trajectories (\S\ref{sec:method}).
    \item We construct a large-scale dataset containing over 500,000 samples from Matterport3D~\cite{chang2017matterport3d} and ScanNet~\cite{dai2017scannet}, including BEV images, instructions, procedurally annotated trajectories, and flow fields (\S\ref{sec:dataset}). 
    \item We benchmark CoFL against modular VLM planners and generative trajectory policies on the proposed dataset with strictly unseen scenes, showing that dense field supervision yields more precise and safer navigation while maintaining real-time performance (\S\ref{sec:experiments}).
    \item We demonstrate zero-shot real-world navigation by deploying a model pretrained on the proposed dataset to a physical robot without real-world adaptation, showcasing the transferability and robustness of the learned flow-field representation (\S\ref{sec:realworld}).
\end{enumerate}

\section{Related Work}
\label{sec:related}

\subsection{Bird's-Eye View Perception for Robotic Navigation}
BEV-aligned representations provide a metric, top-down workspace where geometric constraints and action execution can be expressed globally in a consistent coordinate frame.
Compared to egocentric perspective views, BEV observations reduce scale ambiguity and make free space and obstacle geometry explicit, which is particularly convenient for downstream planning and control.
Despite practical challenges in obtaining BEV images, several settings and recent advances make it increasingly feasible:
(i) In heterogeneous aerial--ground systems, an aerial teammate can provide top-down context to guide ground robots in cluttered or unknown environments~\cite{hsieh2007adaptive, hood2017bird, dang2020graph, queralta2020collaborative, ravichandran2025deploying, liu2025hierarchical}.
(ii) In indoor settings, a calibrated overhead camera (or fixed infrastructure sensing) provides a lightweight way to build metric BEV-aligned workspaces for robot navigation.
(iii) Alternatively, BEV-aligned representations can be constructed onboard from multi-view cameras via view transformation and lifting-based perception, enabling robot-centric BEV-aligned workspaces without external viewpoints~\cite{philion2020lift, li2024bevformer}.

In this work, we instantiate the BEV interface using setting (ii): calibrated top-down and high-oblique observations in experiments, which provide a controlled metric workspace for field-based policy learning.
Its common coordinate frame allows a policy to be queried from many possible starts and enables dense supervision of local guidance over the same scene.

\subsection{Vision-Language-Action Policies}
The intersection of robotics and foundation models has moved beyond discrete, graph-structured VLN benchmarks~\cite{anderson2018vision} toward continuous sensorimotor control. Currently, approaches typically adopt either a hierarchical or an end-to-end paradigm.

\noindent\textbf{Hierarchical instruction following.}
A prevalent line of work leverages LLMs/VLMs to decompose instructions into sub-goals or skill programs.
Systems such as SayCan~\cite{ichter2022saycan} ground language into executable skills via affordance-aware scoring, while other modular pipelines combine semantic grounding with mapping/planning~\cite{shah2022lmnav,gadre2023cows,liang2023code,chen2024mapgpt,liu2024llmhrc}.
While such decomposition improves interpretability and reuses pre-trained priors, it can introduce a semantic--geometric gap: the high-level reasoner may overlook low-level kinematic constraints.

\noindent\textbf{End-to-end Vision-Language-Action/Navigation.}
To unify perception and control, VLA/VLN models learn a direct mapping from observations to actions.
Pioneering works such as RT~\cite{Brohan-RSS-23,zitkovich2023rt} and PaLM-E~\cite{driess2023palm} formulate control as sequence modeling, discretizing the action into categorical tokens.
While scalable, discrete tokenization introduces stepwise actions and struggles with motion smoothness due to the discretized nature of the generated actions.
To mitigate this, recent generative policies formulate the problem as generating continuous action chunks.
Diffusion models~\cite{ho2020denoising,black2024training,chi2025diffusion} treat planning as conditional generation via iterative denoising, achieving strong performance in several manipulation settings.
However, the requirement for multi-step iterative sampling presents a fundamental trade-off between trajectory quality and real-time responsiveness~\cite{song2023consistency}.
Recently, few-step diffusion policy~\cite{clemente2025twosteps} and flow matching~\cite{lipman2023flow, liu2023flow} have emerged as appealing alternatives to multi-step denoising, offering faster inference or straighter probability flow. 
While recent works have successfully applied these frameworks to robot control~\cite{BlackK-RSS-25, black2025pi, intelligence2025pi}, they still instantiate policies as trajectory or action-chunk samplers: given an observation, language instruction, and current state, the model generates a finite-horizon rollout. 
Such policies can be highly expressive, but their supervision is naturally concentrated on the demonstrated starts and the visited states along each trajectory. 
Consequently, workspace structure---for example, how the same instruction should guide motion from other feasible locations in the scene---is only learned indirectly through the trajectory distribution.


\noindent\textbf{Positioning of our work.}
CoFL addresses these limitations by reframing language-conditioned navigation as workspace-level guidance learning.
Motivated by flow matching~\cite{lipman2023flow, liu2023flow} and classical field-based navigation~\cite{khatib1986real,2024NPF}, we formulate the policy as a language-conditioned vector field defined over workspaces.
Compared with action-token or diffusion-style trajectory policies, the key distinction lies in the spatial support and grounding of supervision.
Trajectory generators learn finite vector sequences anchored at demonstrated starts, where later vectors are tied to implicit future states along an unexecuted rollout.
CoFL instead supervises position-conditioned motion vectors across the 2D workspace, so each vector is grounded to a queried physical location.
This changes how each annotation is used: rather than providing supervision only along one trajectory, the same scene--instruction pair constrains a dense set of local control decisions over the workspace.

\section{CoFL Framework}
\label{sec:method}

We present CoFL, a vision-language-action policy that formulates semantic navigation as flow field prediction. This section describes our problem formulation (\S\ref{subsec:problem}), network architecture (\S\ref{subsec:architecture}), training procedure (\S\ref{subsec:training}), and inference strategy (\S\ref{subsec:inference}).
Table~\ref{tab:param_table} summarizes the main symbols used in the framework and procedural annotation pipeline.

\begin{table}[!t]
\centering
\scriptsize
\setlength{\tabcolsep}{2.5pt}
\renewcommand{\arraystretch}{1.08}
\caption{Notation used in this paper.}
\vspace{-2mm}
\label{tab:param_table}
\begin{tabular}{@{}p{0.18\columnwidth} p{0.26\columnwidth} p{0.52\columnwidth}@{}}
	\toprule
    \textbf{Symbol} & \textbf{Domain} & \textbf{Description} \\
\midrule
\multicolumn{3}{c}{\emph{Problem Setup}} \\
\midrule
$\mathcal{E}$ & $[0,1]^2$ & Normalized BEV-aligned workspace. \\
$\mathcal{X}_{\mathrm{free}},\ \mathcal{X}_{\mathrm{obs}}$ & $\subseteq \mathcal{E}$ & Free-space and obstacle subsets. \\
$I$ & $\mathbb{R}^{H\times W\times 3}$ & RGB BEV input (top-down/high-oblique). \\
$\ell$ & text & Language instruction. \\
$\mathbf{x}_0,\ \tau$ & $\mathcal{X}_{\mathrm{free}},\ [0,1]\!\to\!\mathcal{E}$ & Start state and trajectory. \\
$\mathbf{v}_{\theta}(\cdot\mid I,\ell)$ & $\mathcal{E}\!\to\!\mathbb{R}^2$ & Conditioned continuous flow policy. \\
$\mathbf{v}^*(\cdot),\ \mathbf{V}^*$ & $\mathcal{E}\!\to\!\mathbb{R}^2,\ \mathbb{R}^{H\times W\times 2}$ & Supervisory flow field. \\
\midrule
\multicolumn{3}{c}{\emph{Model and Querying}} \\
\midrule
$\mathbf{C}$ & $\mathbb{R}^{N_v\times d}$ & Encoder context tokens. \\
$\mathbf{X}=\{\mathbf{x}_i\}_{i=1}^{N}$ & $[0,1]^{N\times 2}$ & Decoder query coordinates. \\
$\mathbf{M}(\mathbf{X}),\ \mathbf{D}(\mathbf{X})$ & $\mathbb{R}_{>0}^{N\times 1},\ \mathbb{R}^{N\times 2}$ & Predicted speed and direction. \\
$\mathbf{V}(\mathbf{X})$ & $\mathbb{R}^{N\times 2}$ & Queried velocity vectors. \\
\midrule
\multicolumn{3}{c}{\emph{Training and inference}} \\
\midrule
$N_s,\ g,\ N_{\mathrm{bin}}$ & $\mathbb{Z}_+$ & Training queries, grid size, and per-cell count. \\
$\lambda,\ \epsilon$ & $\mathbb{R}_+$ & Loss weight and numerical stabilizer. \\
$\tilde{g},\ \hat{\mathbf{V}}$ & $\mathbb{Z}_+,\ \mathbb{R}^{\tilde{g}\times\tilde{g}\times 2}$ & Inference grid and queried dense flow. \\
$T,\ \Delta t,\ t_k$ & $\mathbb{Z}_+,\ \mathbb{R}_+,\ [0,1]$ & Rollout horizon, step size, and time index. \\
\midrule
\multicolumn{3}{c}{\emph{Annotation Pipeline}} \\
\midrule
$\mathcal{S}$ & $\mathbb{N}^{H\times W}$ & Semantic BEV label map. \\
$\mathcal{R}$ & labels $\to$ roles & Traversability/target label map. \\
$\ell_{\mathrm{target}},\ \mathbf{x}_g$ & text,\ $\mathcal{X}_{\mathrm{free}}$ & Target phrase and goal sources. \\
$D_g^w,\ D_{\mathrm{obs}},\ \Phi$ & $\mathcal{X}_{\mathrm{free}}\!\to\!\mathbb{R}_+$, $\mathcal{X}_{\mathrm{obs}}\!\to\!\mathbb{R}_+$,\ $\mathcal{E}\!\to\!\mathbb{R}_+$ & Cost-weighted geodesic distance, distance-to-free, and potential. \\
\bottomrule
\end{tabular}
\end{table}

\subsection{Problem Formulation}
\label{subsec:problem}

Consider a mobile robot operating in a BEV-aligned workspace $\mathcal{E}=[0,1]^2$ (using normalized image coordinates: $x$ right, $y$ down) with free space $\mathcal{X}_{\text{free}}\subset\mathcal{E}$ and obstacle regions $\mathcal{X}_{\text{obs}}=\mathcal{E}\setminus\mathcal{X}_{\text{free}}$. Given a top-down or high-oblique RGB observation $I \in \mathbb{R}^{H \times W \times 3}$ and a natural language instruction $\ell$ specifying a target object (e.g., ``navigate to the sofa''), our objective is to learn a policy that generates collision-free paths to the goal from arbitrary initial configurations.

We depart from conventional formulations that predict discrete action sequences or waypoint trajectories. Instead, we model navigation as learning a goal-conditioned flow field
\begin{equation}
\mathbf{v}_\theta: \mathcal{E} \times \mathcal{I} \times \mathcal{L} \rightarrow \mathbb{R}^2,
\label{eq:velocity_field}
\end{equation}
where $\theta$ denotes learnable parameters, $\mathcal{I}$ is the space of observations, and $\mathcal{L}$ is the space of language instructions. 

We parameterize a planar trajectory $\tau(t)\in\mathcal{E}$ by normalized time $t\in[0,1]$.
The rollout is obtained by integrating the predicted flow field $\mathbf{v}\bigl(\cdot)$:
\begin{equation}
\dot{\tau}(t) = \mathbf{v}\bigl(\tau(t)\mid I,\ell\bigr),\quad \tau(0)=\mathbf{x}_0,
\label{eq:ode}
\end{equation}
where $\mathbf{x}_0\in\mathcal{X}_{\mathrm{free}}$.
We aim for collision-free motion by encoding obstacle awareness in the supervision $\mathbf{v}^*$
(geodesic-to-goal guidance with repulsion, detailed in \S\ref{sec:dataset}), which steers the rollout away from $\mathcal{X}_{\mathrm{obs}}$.


\begin{figure*}[t]
  \centering
  \includegraphics[width=0.95\linewidth]{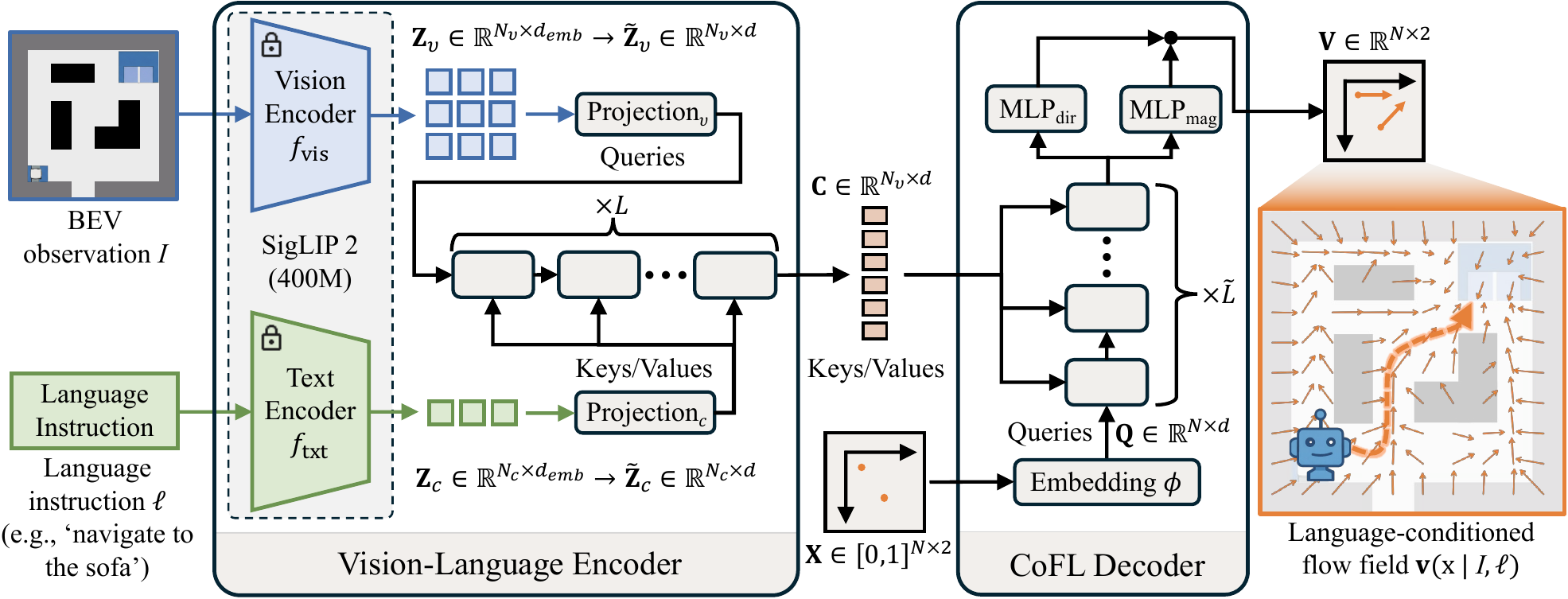}
  \caption{{Overview of the CoFL's network architecture.} Given an RGB BEV observation $I$ and a language instruction $\ell$, a SigLIP 2-based~\cite{zhai2023sigmoid,tschannen2025siglip2} vision--language encoder produces language-conditioned context tokens over the BEV image. The decoder then queries this context with 2D normalized spatial coordinates $\mathbf{X}$ and outputs the corresponding velocities $\mathbf{V}$, forming a continuous flow field $\mathbf{v}(\mathbf{x} \mid I, \ell)$ via bilinear interpolation of discrete velocity predictions.}
  \label{fig:network}
  \vspace{-3mm}
\end{figure*}

\subsection{Network Architecture}
\label{subsec:architecture}

The CoFL architecture comprises two modules: a vision-language encoder $f_{\text{enc}}$ that extracts goal-conditioned scene representations and a decoder $f_{\text{flow}}$ that maps spatial coordinates to velocity vectors (Fig.~\ref{fig:network}).

\subsubsection{Vision-Language Encoder}
\label{subsec:encoder}
We leverage SigLIP 2~\cite{zhai2023sigmoid,tschannen2025siglip2} as the frozen backbone ($f_{\text{vis}},f_{\text{txt}}$) for joint vision-language embedding. Given observation $I$ and instruction $\ell$, we extract
\begin{align}
\mathbf{Z}_v &= f_{\text{vis}}(I) \in \mathbb{R}^{N_v \times d_{\text{emb}}}, \\
\mathbf{Z}_c &= f_{\text{txt}}(\ell) \in \mathbb{R}^{N_c \times d_{\text{emb}}},
\end{align}
where $N_v$ is the number of visual tokens, $N_c$ is the number of text tokens, including a prepended global token, and $d_{\text{emb}}$ is the embedding dimension. 

We first project visual and text tokens into a shared hidden dimension $d$ using a linear projection followed by LayerNorm.
\begin{align}
\tilde{\mathbf{Z}}_v &= \text{LN}\big(\text{Projection}_v(\mathbf{Z}_v)\big) \in \mathbb{R}^{N_v \times d}, \\
\tilde{\mathbf{Z}}_c &= \text{LN}\big(\text{Projection}_c(\mathbf{Z}_c)\big) \in \mathbb{R}^{N_c \times d}.
\end{align}

These projected tokens are then fused using $L$ transformer decoder layers~\cite{ashish2017attn, carion2020detr}. Starting from $\mathbf{H}^{(0)} = \tilde{\mathbf{Z}}_v$, we apply
\begin{equation}
\mathbf{H}^{(l)} =\mathcal{D}^{(l)}\bigl(\mathbf{H}^{(l-1)},\, \tilde{\mathbf{Z}}_c\bigr), \quad l \in \{1 \dots L\},
\label{eq:vl_decoder}
\end{equation}
where each decoder layer follows a standard transformer decoder block $\mathcal{D}$, consisting of self-attention (non-causal), cross-attention (with keys/values from ${\tilde{\mathbf{Z}}_c}$), and an FFN, each wrapped with residual connections and normalization.
\begin{equation}
\mathbf{C} = \mathbf{H}^{(L)} \in \mathbb{R}^{N_v \times d},
\end{equation}
which provides a task-relevant understanding of the environment.

\subsubsection{CoFL Decoder}
The decoder adopts a query-based design. 
Given $N$ spatial coordinates $\mathbf{X} \in [0,1]^{N\times 2}$ (normalized to the image frame), where the $i$-th row is $\mathbf{x}_i \in [0,1]^2$, we first compute positional embeddings and project them to $d$-dimensional query tokens:
\begin{equation}
\mathbf{Q} = \phi_{\text{pos}}(\mathbf{X}) \in \mathbb{R}^{N \times d},
\label{eq:pos_embed}
\end{equation}
where $\phi_{\text{pos}}$ is implemented as a Gaussian Fourier encoding~\cite{tancik2020fourier} followed by a linear projection and LayerNorm, providing a rich continuous representation for coordinate queries.

The query tokens then attend to the context tokens through $\tilde{L}$ simplified transformer decoder layers~\cite{ashish2017attn,carion2020detr}:
\begin{equation}
\tilde{\mathbf{H}}^{(\tilde{l})} = \tilde{\mathcal{D}}^{(\tilde{l})}\bigl(\tilde{\mathbf{H}}^{(\tilde{l}-1)}, \mathbf{C}\bigr), 
\quad \tilde{l} \in \{1,\dots,\tilde{L}\},
\label{eq:decoder}
\end{equation}
with $\tilde{\mathbf{H}}^{(0)} = \mathbf{Q}$. Each decoder layer follows a simplified transformer decoder block $\tilde{\mathcal{D}}$ without query self-attention: it performs cross-attention from $\tilde{\mathbf{H}}^{(\tilde{l}-1)}$ (queries) to the context tokens $\mathbf{C}$ (keys/values), followed by an FFN; both sublayers are wrapped with residual connections and normalization.

Instead of regressing $(v_x,v_y)$ directly, we predict a positive magnitude and a unit direction for each query:
\begin{align}
\mathbf{M}(\mathbf{X}) &= \text{Softplus}\!\left(\text{MLP}_{\text{mag}}(\tilde{\mathbf{H}}^{(\tilde{L})})\right) \in \mathbb{R}_{> 0}^{N \times 1}, 
\label{eq:mag_head}\\
\mathbf{D}(\mathbf{X}) &= \frac{\text{MLP}_{\text{dir}}(\tilde{\mathbf{H}}^{(\tilde{L})})}{\left\|\text{MLP}_{\text{dir}}(\tilde{\mathbf{H}}^{(\tilde{L})})\right\|_2 + \epsilon} \in \mathbb{R}^{N \times 2},
\label{eq:dir_head}
\end{align}
and compose the final velocity as
\begin{equation}
\mathbf{V}(\mathbf{X}) = \mathbf{M}(\mathbf{X}) \odot \mathbf{D}(\mathbf{X}) \in \mathbb{R}^{N \times 2},
\label{eq:velocity_head}
\end{equation}
where $\odot$ denotes broadcasting element-wise multiplication and $\epsilon$ is a small constant for numerical stability.
We interpret $\mathbf{M}(\mathbf{X})$ as the local motion magnitude implied by the supervision, and $\mathbf{D}(\mathbf{X})$ as the corresponding local motion direction.

\subsection{Training}
\label{subsec:training}

We train CoFL by supervising the predicted flow at $N_s$ query locations per training instance.
Given queries $\mathbf{x}_i\in[0,1]^2$, we optimize $\mathbf{v}(\mathbf{x}_i)$ as a continuous-coordinate notation for the discrete samples $\mathbf{V} \in \mathbb{R}^{N_s\times 2}$, targets $\mathbf{v}^*(\mathbf{x}_i)$ are obtained by bilinearly sampling the annotated field
$\mathbf{V}^*\in\mathbb{R}^{H\times W\times 2}$ (See \S\ref{sec:dataset} for details of annotation).

\subsubsection{Efficient Sampling}
Dense supervision over all pixels is expensive. Instead, the $N_s$ query locations are constructed via stratified sampling.
We partition $\mathcal{E}=[0,1]^2$ into a $g\times g$ grid of cells $\{\mathcal{G}_j\}_{j=1}^{g^2}$ and draw $N_{\mathrm{bin}}$ jittered points uniformly within each cell:
\begin{equation}
\mathbf{x}_{j,r} \sim \mathrm{U}(\mathcal{G}_j),\quad r=1,\dots,N_{\mathrm{bin}},
\label{eq:stratified_sampling}
\end{equation}
where $N_{\mathrm{bin}}=\lceil N_s/g^2\rceil$.
We concatenate all sampled points and keep $N_s$ of them to form $\mathbf{X}=\{\mathbf{x}_i\}_{i=1}^{N_s}$.

\subsubsection{Loss Function}
Given the sampled queries $\mathbf{X}\sim p(\mathbf{X})$, the objective combines direction and magnitude supervision:
\begin{equation}
\mathcal{L} = \mathcal{L}_{\text{dir}} + \lambda \mathcal{L}_{\text{mag}}.
\label{eq:total_loss}
\end{equation}
The direction loss enforces angular alignment via cosine similarity:
\begin{equation}
\mathcal{L}_{\text{dir}} =
\mathbb{E}_{\mathbf{X}}
\left[\frac{1}{N_s}\sum_{i=1}^{N_s}
\left(1-\frac{\mathbf{v}(\mathbf{x}_i)^\top \mathbf{v}^*(\mathbf{x}_i)}
{\|\mathbf{v}(\mathbf{x}_i)\|_2\,\|\mathbf{v}^*(\mathbf{x}_i)\|_2+\epsilon}\right)\right],
\label{eq:dir_loss}
\end{equation}
and the magnitude loss matches velocity norms:
\begin{equation}
\mathcal{L}_{\text{mag}}=
\mathbb{E}_{\mathbf{X}}
\left[\frac{1}{N_s}\sum_{i=1}^{N_s}
\left(\|\mathbf{v}(\mathbf{x}_i)\|_2-\|\mathbf{v}^*(\mathbf{x}_i)\|_2\right)^2\right].
\label{eq:mag_loss}
\end{equation}

\subsection{Inference}
\label{subsec:inference}

\begin{figure}[t]
  \centering
  \includegraphics[width=0.95\linewidth]{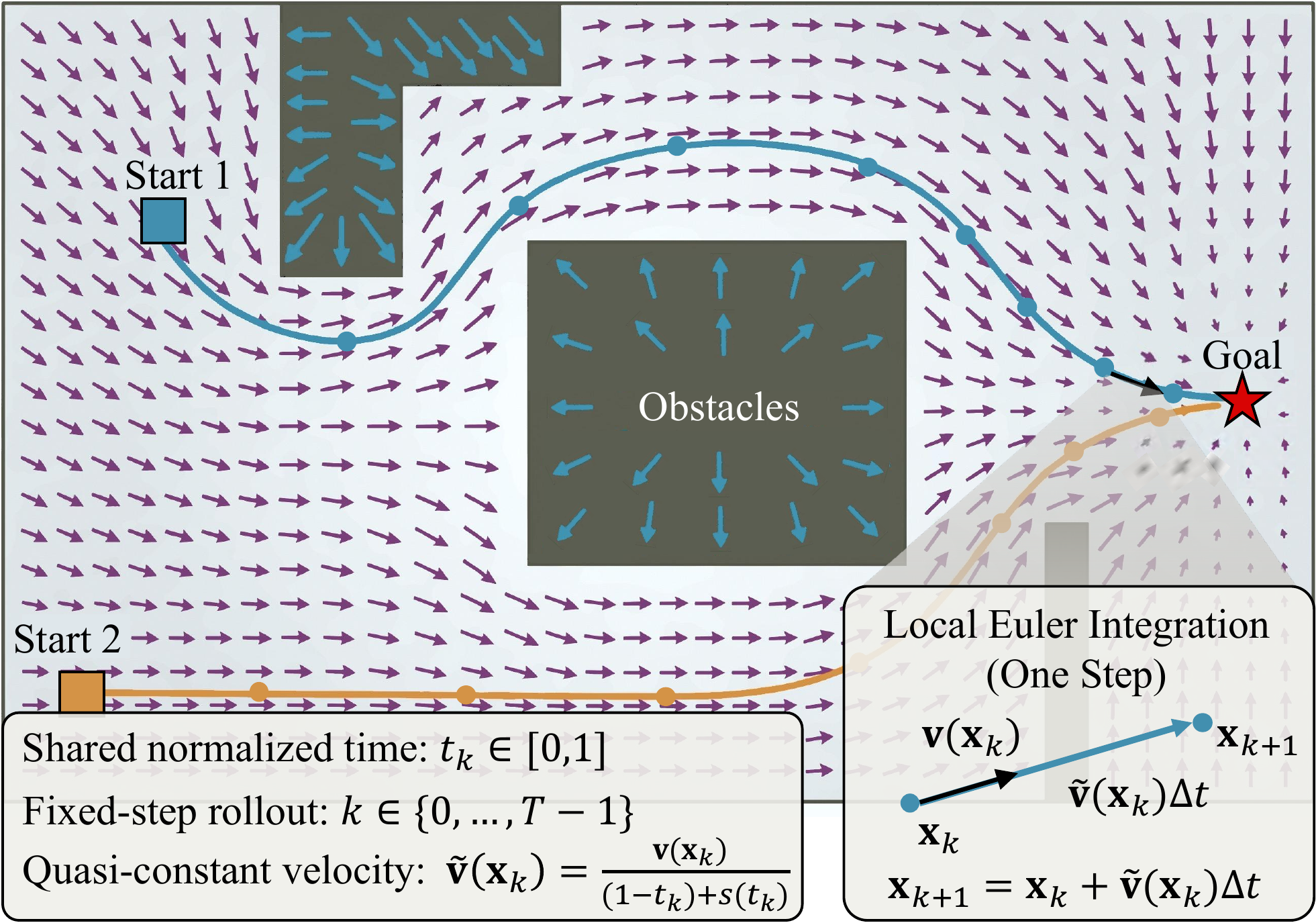}
  \caption{{Overview of the trajectory inference.} The predicted flow field over the workspace guides agents from different starts toward the same goal while smoothly avoiding obstacles.}
  \label{fig:infer}
\end{figure}

We generate trajectories by numerically integrating the predicted flow field, as illustrated in Fig.~\ref{fig:infer}.
For efficiency, we first query the decoder on a $\tilde{g}\times\tilde{g}$ coordinate grid to obtain a dense flow grid
$\hat{\mathbf{V}}\in\mathbb{R}^{\tilde{g}\times\tilde{g}\times 2}$.
Given a start position $\mathbf{x}_0$ and a horizon of $T$ steps, we perform forward Euler integration with $\Delta t=1/T$ and $t_k=k/T$.
At step $k\in\{0,\dots,T-1\}$, we obtain $\mathbf{v}(\mathbf{x}_k \mid I,\ell)$ by bilinearly sampling $\hat{\mathbf{V}}$ at $\mathbf{x}_k$.

In our procedural annotations (\S\ref{sec:dataset}), the field magnitude is constructed to correlate with the remaining normalized distance-to-go (which decays as $t\!\rightarrow\!1$). Thus, to obtain a quasi-constant-velocity rollout toward the goal, we apply a deterministic inverse-time rescaling:
\begin{equation}
\tilde{\mathbf{v}}_k 
=
\frac{\mathbf{v}(\mathbf{x}_k \mid I,\ell)}{(1-t_k)+\beta\,t_k^{\alpha}},
\qquad t_k \in [0,1],
\label{eq:inv_schedule}
\end{equation}
where $\alpha$ and $\beta$ are stabilizer parameters. The ``soft cap'' prevents the $1/(1-t)$ factor from exploding as $t\to 1$.

We then update the state by:
\begin{equation}
\mathbf{x}_{k+1} = \mathbf{x}_k + \tilde{\mathbf{v}}_k\Delta t, \quad \mathbf{x}_{k+1} \leftarrow \text{clip}(\mathbf{x}_{k+1}, 0, 1).
\label{eq:euler}
\end{equation}

By predicting both direction and magnitude, trajectories $\tau$ can be generated from arbitrary $\mathbf{x}_0$ over $t_k\in[0,1)$.

\section{Large-Scale BEV Navigation Dataset Construction}
\label{sec:dataset}
To evaluate our CoFL, we constructed a dataset focusing on language-conditioned navigation. Formally, our dataset consists of $N$ samples $\mathbb{D} = \{(I_i, \ell_i, \mathbf{V}^*_i, \tau^*_i)\}_{i=1}^N$, where $I$ represents the RGB BEV observation, $\ell$ is the instruction, and $\mathbf{V}^*$ and $\tau^*$ denote procedurally generated flow field and trajectory annotations. The pipeline is detailed below.

\subsection{Visual Observation Generation}
\label{subsec:image_collection}
\begin{figure}[t]
  \centering
  \includegraphics[width=0.9\linewidth]{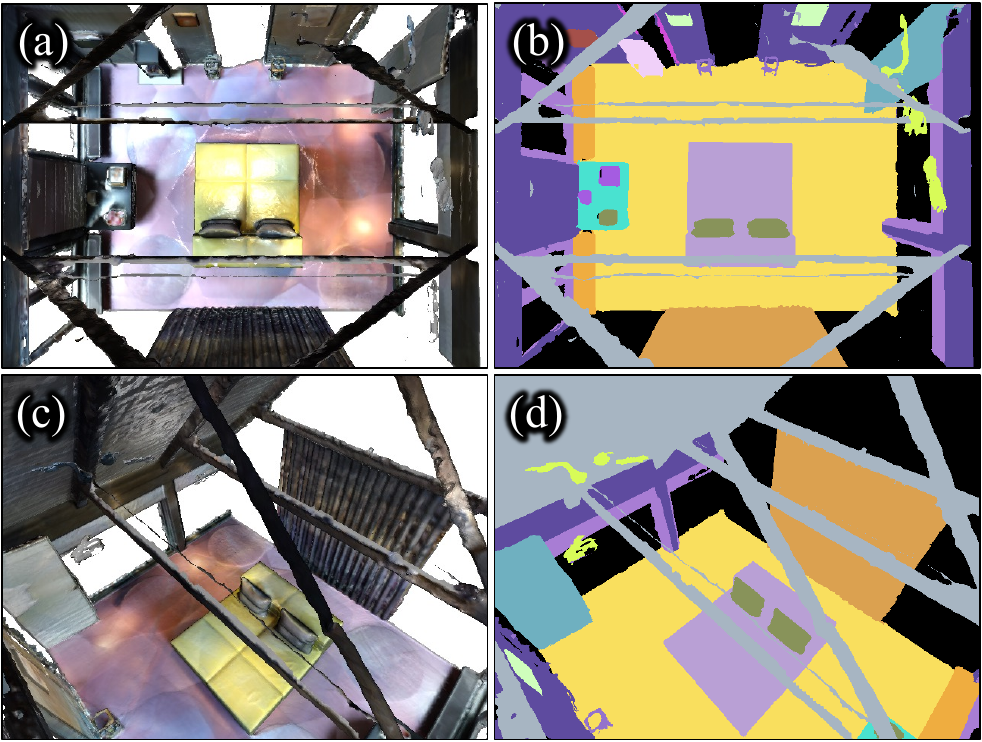}
  \caption{{Overview of the proposed visual observation generation pipeline.} Images are captured from a multi-view camera array. (a)-(b) Top-down view RGB and semantic segmentation maps. (c)-(d) Oblique view sample.}
  \label{fig:image_collection}
  \vspace{-3mm}
\end{figure}

We utilize high-quality scenes from two large-scale 3D indoor scene libraries: \textbf{Matterport3D}~\cite{chang2017matterport3d} and \textbf{ScanNet}~\cite{dai2017scannet}. As detailed in Fig.~\ref{fig:image_collection} and Table~\ref{tab:datasets}, for each region, we generate 1 top-view and 8 multi-oblique views with randomized pitch angles ($45^\circ$-$80^\circ$). Specifically, we render these RGB images as observations $I$ and corresponding annotated semantic maps $\mathcal{S}$. The semantic maps are used to instantiate language instructions $\ell$ and to derive procedural flow field and trajectory annotations ($\mathbf{V}^*$,$\tau^*$).
For reproducibility, we also record per-view camera calibrations.

\begin{table}[!t]
\centering
\caption{Statistics of the Implemented Datasets.}
\vspace{-1mm}
\label{tab:datasets}
\resizebox{\linewidth}{!}{%
\begin{tabular}{lcccc}
\toprule
Dataset & \# Scenes & \# Regions & \# Views & \# Samples \\
\midrule
Matterport3D~\cite{chang2017matterport3d} & 90 & 2,056 & \multirow{2}{*}{9} & 245,604\\
ScanNet~\cite{dai2017scannet} & 1,513 & 1,513  & & 342,038\\
\bottomrule
\end{tabular}%
}
\vspace{-2mm}
\end{table}

\subsection{Language Instruction Generation} 
\label{subsec:instruction_collection}
To enable language conditioning, we synthesize instructions using diverse, multi-template patterns. 
We combine action verbs (e.g., \textit{``Navigate to''}, \textit{``move toward''}) with a target phrase $\ell_{\mathrm{target}}$, formed by the target name with optional spatial modifiers (e.g., \textit{``left of''}, \textit{``behind''}) and optional relative descriptors when needed (e.g., \textit{``the second from the upper side''}).
We define a dataset-specific label mapping $\mathcal{R}$ based on the official taxonomy, partitioning labels into free space and obstacles; the target name in $\ell_{\mathrm{target}}$ is drawn from a targetable subset of obstacle categories in $\mathcal{R}$ for instruction synthesis.

\subsection{Procedural Annotation}
\label{subsec:vel_field_generation}

\begin{figure}[!t]
  \centering
  \includegraphics[width=0.73\linewidth]{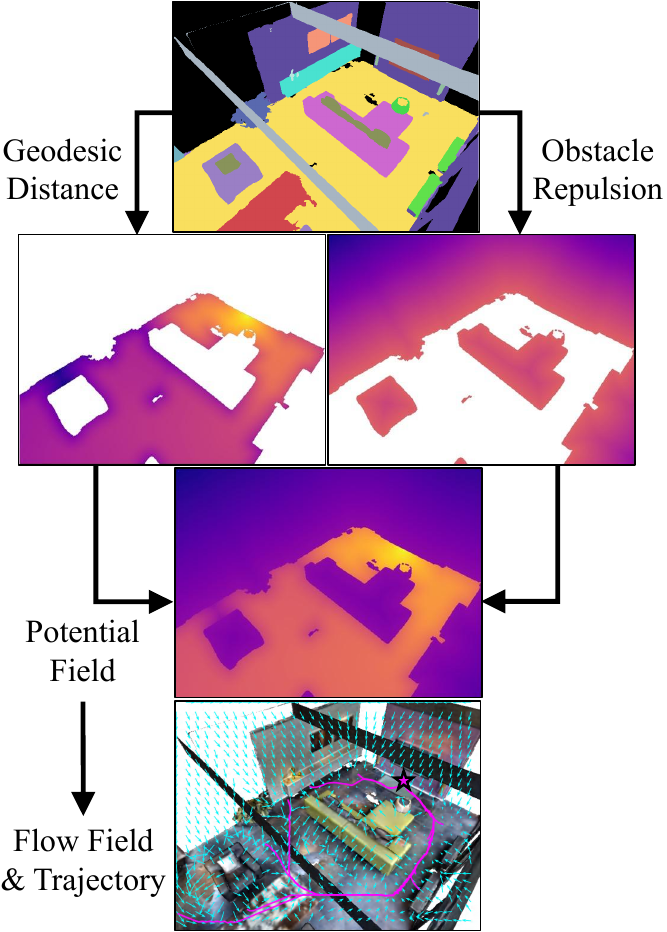}
  \caption{{Overview of the procedural annotation pipeline.} We combine cost-weighted geodesic distance and obstacle repulsion to derive the field and trajectory annotations.}
  \label{fig:gt_pipeline}
\end{figure}

We derive flow fields and trajectories using the semantic maps from \S\ref{subsec:image_collection} and the targets defined in \S\ref{subsec:instruction_collection}. 
For each (map, target) pair, the flow field and trajectory are annotated by the pipeline in Algorithm~\ref{alg:gt_generation} and illustrated in Fig.~\ref{fig:gt_pipeline}.
The key idea is to construct a goal-attractive and obstacle-aware potential field by combining cost-weighted geodesic distance with obstacle repulsion.

\begin{algorithm}[!t]
\caption{Procedural Annotation Pipeline}
\label{alg:gt_generation}
\small
\begin{algorithmic}[1]
\Require semantic map $\mathcal{S}$, label mapping $\mathcal{R}$, and target $\ell_{\mathrm{target}}$
\Ensure flow field $\mathbf{V}^*$ and trajectory $\tau^*$
\vspace{0.1cm}
\hrule
\vspace{0.1cm}

\Statex \textbf{\textcolor{blue!70!black}{Stage 1: Traversability \& Goal Extraction}}
\State $\mathcal{M}_{\mathrm{free}} \gets \textsc{ExtractFree}(\mathcal{S},\mathcal{R})$ \Comment{free: 1, obstacle: 0}
\State $\mathcal{M}_{\mathrm{obs}} \gets \neg \mathcal{M}_{\mathrm{free}}$
\State $\mathbf{x}_g \gets \textsc{ComputeGoal}(\mathcal{S}, \ell_{\mathrm{target}})$ \Comment{goal sources from map}
\Statex \textbf{\textcolor{blue!70!black}{Stage 2: Cost-Weighted Geodesic Distance}}
\State $D_{\mathrm{free}} \gets \textsc{DTO}(\mathcal{M}_{\mathrm{free}})$ \Comment{distance-to-obstacle in free space}
\State ${C}_\text{cost} \gets \textsc{CostMap}(D_{\mathrm{free}}, \lambda_{\mathrm{safe}}, \rho_{\mathrm{safe}})$
\Comment{truncated linear penalty if $D_{\mathrm{free}}<\rho_{\mathrm{safe}}$; $\lambda_{\mathrm{safe}}$ is the penalty coefficient}

\State $(D_g^w,\; \text{pred}) \gets \textsc{Geodesic}(\mathcal{M}_{\mathrm{free}}, {C}_\text{cost}, \mathbf{x}_g)$\Comment{cost-weighted distance and predecessor map (Dijkstra~\cite{dljkstra1959note})}
\State $D_g^{\mathrm{pix}} \gets \textsc{PixelLengthFromPred}(\text{pred})$ 
\Comment{pixel length of path}
\Statex \textbf{\textcolor{blue!70!black}{Stage 3: Obstacle Inner Repulsion}}
\State $D_{\mathrm{obs}} \gets \textsc{DTF}(\mathcal{M}_{\mathrm{obs}})$ \Comment{distance-to-free in obstacle space}
\Statex \textbf{\textcolor{blue!70!black}{Stage 4: Potential Field Construction}}
\State $\Phi(\mathbf{x}) \gets
\begin{cases}
w_g D_g^w(\mathbf{x}), & \mathbf{x}\in \mathcal{M}_{\mathrm{free}}\\
w_{\mathrm{obs}}D_{\mathrm{obs}}(\mathbf{x})+b_{\mathrm{obs}}, & \mathbf{x}\in \mathcal{M}_{\mathrm{obs}}
\end{cases}$
\Comment{$w_\mathrm{obs} \gg w_g$; $b_{\mathrm{obs}}$ shifts obstacle potential upward at the boundary}
\Statex \textbf{\textcolor{blue!70!black}{Stage 5: Flow Field Generation}}
\State $\mathbf{u}(\mathbf{x}) \gets -\nabla \Phi(\mathbf{x}) / (\|\nabla \Phi(\mathbf{x})\|+\epsilon)$
\If{$\mathcal{M}_{\mathrm{free}}(\mathbf{x})=1$}
   \State $\mathbf{V}^*(\mathbf{x}) \gets [u_x D_g^{\mathrm{pix}}/W,\; u_y D_g^{\mathrm{pix}}/H]$ \Comment{distance-to-go scaling}
\Else
   \State $\mathbf{V}^*(\mathbf{x}) \gets \mathbf{u}(\mathbf{x})$ \Comment{unit-speed escape}
\EndIf
\Statex \textbf{\textcolor{blue!70!black}{Stage 6:  Trajectory Extraction}}
\State $\mathbf{x}_0 \gets \textsc{SampleStart}(\mathcal{M}_{\mathrm{free}}, D_g^{\mathrm{pix}})$
\Comment{sample reachable start}
\State $\tau_{\mathrm{raw}} \gets \textsc{BacktrackPred}(\text{pred}, \mathbf{x}_0)$
\Comment{follow pred to goal}
\State $\tau^* \gets \textsc{Resample}(\tau_{\mathrm{raw}})$
\Comment{fixed-length resampling}

\State \Return $(\mathbf{V}^*,\,\tau^*)$
\end{algorithmic}
\end{algorithm}

The annotation procedure consists of six stages:
\begin{itemize}
    \setlength\itemsep{0.08em}

    \item \textbf{Stage 1: Traversability and goal extraction.}
    We convert the semantic map $\mathcal{S}$ into a binary free-space mask $\mathcal{M}_{\mathrm{free}}$ and an obstacle mask $\mathcal{M}_{\mathrm{obs}}=\neg\mathcal{M}_{\mathrm{free}}$.
    Goal sources $\mathbf{x}_g$ are extracted from the target region specified by $\ell_{\mathrm{target}}$.

    \item \textbf{Stage 2: Cost-weighted geodesic distance.}
    To encourage safe navigation, we compute a distance-to-obstacle field $D_{\mathrm{free}}$ over free space and convert it into a safety-aware traversal cost:
    \begin{equation}
        C_{\mathrm{cost}}(\mathbf{x})
        =
        1+\lambda_{\mathrm{safe}}
        [\rho_{\mathrm{safe}}-D_{\mathrm{free}}(\mathbf{x})]_+ ,
        \label{eq:main_costmap}
    \end{equation}
    where $[z]_+=\max(0,z)$.
    Thus, cells closer than the safety radius $\rho_{\mathrm{safe}}$ receive larger traversal costs.
    A multi-source Dijkstra \cite{dljkstra1959note} search from the goal sources then produces the cost-weighted distance-to-go field $D_g^w$ and the predecessor map $\mathrm{pred}$.

    \item \textbf{Stage 3: Obstacle inner repulsion.}
    We compute an obstacle-space distance transform $D_{\mathrm{obs}}$, which is positive inside obstacles and measures the distance to the nearest free-space cell.
    This term provides outward gradients for invalid states that fall inside obstacles.

    \item \textbf{Stage 4: Potential field construction.}
    We define a piecewise potential field:
    \begin{equation}
        \Phi(\mathbf{x}) =
        \begin{cases}
        w_g D_g^w(\mathbf{x}), 
        & \mathbf{x}\in \mathcal{M}_{\mathrm{free}},\\
        w_{\mathrm{obs}}D_{\mathrm{obs}}(\mathbf{x})+b_{\mathrm{obs}},
        & \mathbf{x}\in \mathcal{M}_{\mathrm{obs}},
        \end{cases}
        \label{eq:main_potential}
    \end{equation}
    where $w_{\mathrm{obs}}\gg w_g$ and $b_{\mathrm{obs}} = \max_{\mathbf{x}\,:\,\mathcal{M}_{\mathrm{free}}(\mathbf{x})=1} w_g D_g^w(\mathbf{x})$.
    The free-space component attracts states toward the target along low-cost geodesic routes, while the obstacle component pushes states outward from obstacle interiors.

    \item \textbf{Stage 5: Flow-field generation.}
    The ground-truth flow direction is obtained from the negative potential gradient:
    \begin{equation}
        \mathbf{u}(\mathbf{x})
        =
        -\frac{\nabla \Phi(\mathbf{x})}
        {\|\nabla \Phi(\mathbf{x})\|+\epsilon}.
        \label{eq:main_flow_dir}
    \end{equation}
    In free space $\mathbf{x}\in\mathcal{M}_{\mathrm{free}}$, we scale the vector magnitude by the remaining path length $D_g^{\mathrm{pix}}$:
    \begin{equation}
        \mathbf{V}^*(\mathbf{x})
        =
        \left[
        u_x(\mathbf{x})\frac{D_g^{\mathrm{pix}}(\mathbf{x})}{W},
        \;
        u_y(\mathbf{x})\frac{D_g^{\mathrm{pix}}(\mathbf{x})}{H}
        \right].
        \label{eq:main_flow_scale}
    \end{equation}
    Therefore, each free-space vector encodes both the local descent direction and an approximate normalized distance-to-go.
    Inside obstacles $\mathbf{x}\in\mathcal{M}_{\mathrm{obs}}$, we use a unit-speed escape vector $\mathbf{V}^*(\mathbf{x})=\mathbf{u}(\mathbf{x})$.

    \item \textbf{Stage 6: Trajectory extraction.}
    We sample reachable start states, backtrack the predecessor map $\mathrm{pred}$ to obtain cost-least polylines toward the goal, and resample them into fixed-length reference trajectories $\tau^*$.
\end{itemize}

More implementation details of the subroutines in Algorithm~\ref{alg:gt_generation}, including pixel-grid to normalized-coordinate conversion, distance transforms, edge-cost computation, potential smoothing, gradient computation, and trajectory resampling, are provided in Appendix~\ref{app:gt_flow}.

\section{Benchmark Setup and Results}
We evaluate CoFL on language-conditioned navigation across diverse indoor scenes annotated in \S\ref{sec:dataset}. 
This section addresses three questions: 
(1) How well does CoFL predict fields and generalize to unseen scenes (\S\ref{subsec:results})?  
(2) How does CoFL compare to modular or generative baselines in this task (\S\ref{subsec:results})? 
(3) What accounts for CoFL's advantage, and how important is dense workspace-level supervision (\S\ref{subsec:ablation})?
\label{sec:experiments}
\subsection{Experimental Setup}
\label{subsec:setup}

\subsubsection{Implementation Details}
\begin{figure}[!b]
  \centering
  \includegraphics[width=\linewidth]{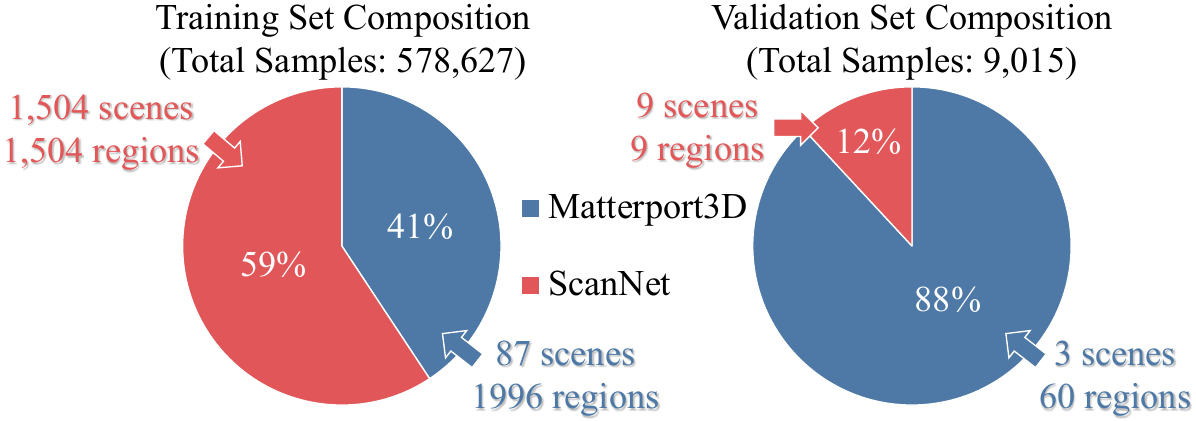}
  \caption{{Overview of the training/validation split.} We employ a strict scene-wise split to assess generalization.}
  \label{fig:dataset_split}
  \vspace{-2mm}
\end{figure}

CoFL uses a frozen SigLIP-2 Base (ViT-B/16)~\cite{zhai2023sigmoid,tschannen2025siglip2} backbone with 4 additional layers as the encoder, and a 2-layer decoder ($d{=}768$).
We train it for 50 epochs with AdamW (lr $10^{-4}$, wd $10^{-5}$) on an RTX 4090 (batch 32; 1,000 samples/training step) and run 100-step Euler rollout at inference.
We adopt a strict scene-wise split on Matterport3D/ScanNet with disjoint validation scenes (Fig.~\ref{fig:dataset_split}).

\begin{figure*}[!t]
  \centering
  \includegraphics[width=\linewidth]{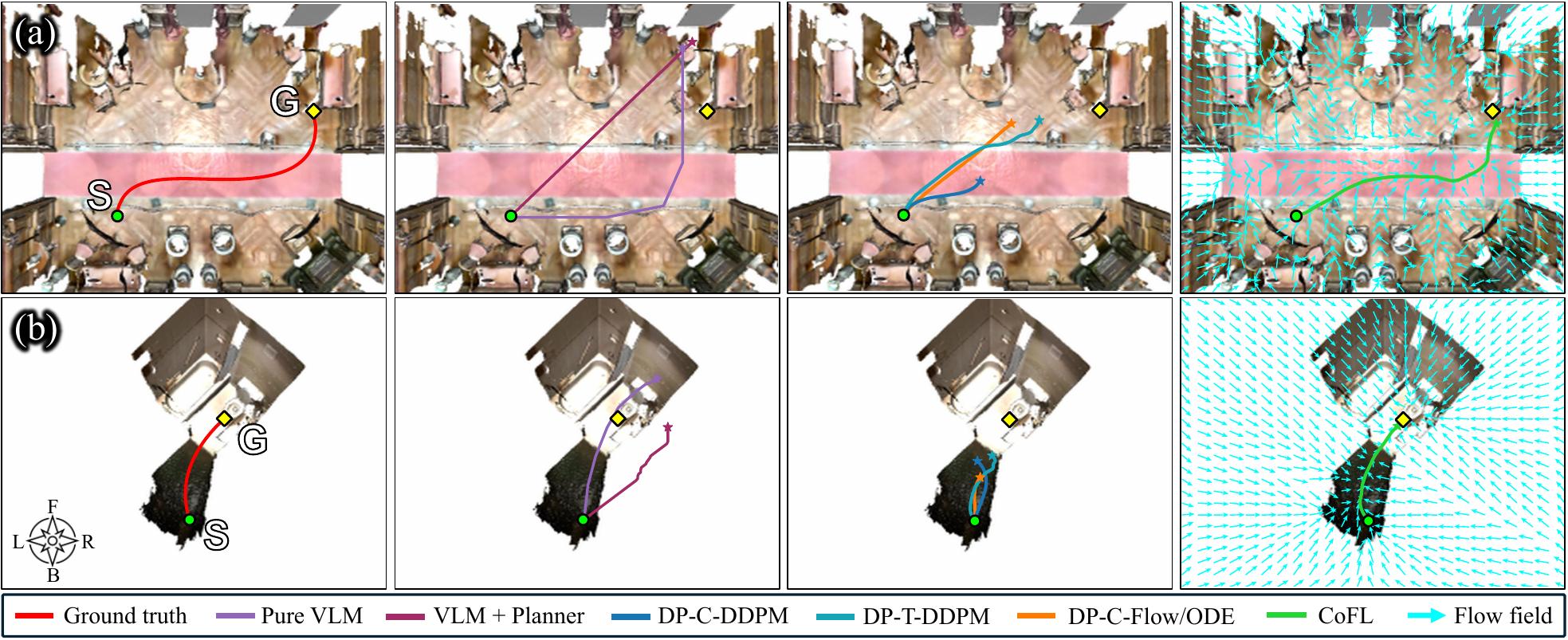}
  \vspace{-5mm}
  \caption{
  {Examples of trajectories on validation set.}
  (a) Matterport3D: \textit{``Please navigate to the sofa in the upper right"}; (b) ScanNet: \textit{``Please go to the toilet"}. From left to right: ground truth, pure VLM/VLM+Planner, DP-family and CoFL. More examples are provided in Appendix~\ref{app:more_example_benchmark}.
  }
  \label{fig:comparision}
  \vspace{-2mm}
\end{figure*}

\subsubsection{Evaluation Metrics}
\label{subsec:metrics}
We evaluate both the predicted trajectories and flow fields.
For trajectories, we report \textbf{FGE} (final goal error; normalized distance between the end of the trajectory and the annotated trajectory endpoint), \textbf{CR} (collision rate; fraction of episodes with any collision), \textbf{Curv} (curvature; mean turning angle in radians computed from successive segments), and \textbf{PLR} (path length ratio; ratio to the path length of the annotated trajectory).
For fields, we report \textbf{AE} (mean angular error; degrees) and \textbf{ME} (mean magnitude error; normalized norm difference).
Details of the calculation are provided in Appendix~\ref{app:metrics}.

\subsubsection{Baselines}
\label{subsec:baselines}
We compare CoFL with representative language-conditioned navigation baselines.
All methods are evaluated in an open-loop setting from the same inputs $(I,\ell,\mathbf{x}_0)$.
For a fair comparison within the DP family, all DP variants reuse the same vision--language encoder as CoFL (\S\ref{subsec:encoder}) and differ only in the trajectory generator.

\noindent\textbf{VLM-based navigation.}
(i) \emph{Pure VLM}: a VLM directly predicts a waypoint sequence~\cite{pmlr-v288-goetting25a};
(ii) \emph{VLM+Planner}: a VLM predicts the goal and obstacles followed by a geometric planner~\cite{liu2025hierarchical}. Gemini-2.5-Flash (API) is used as the VLM.

\begin{figure*}[!t]
  \centering
  \begin{subfigure}[b]{0.24\linewidth}
        \centering
        \includegraphics[width=\linewidth]{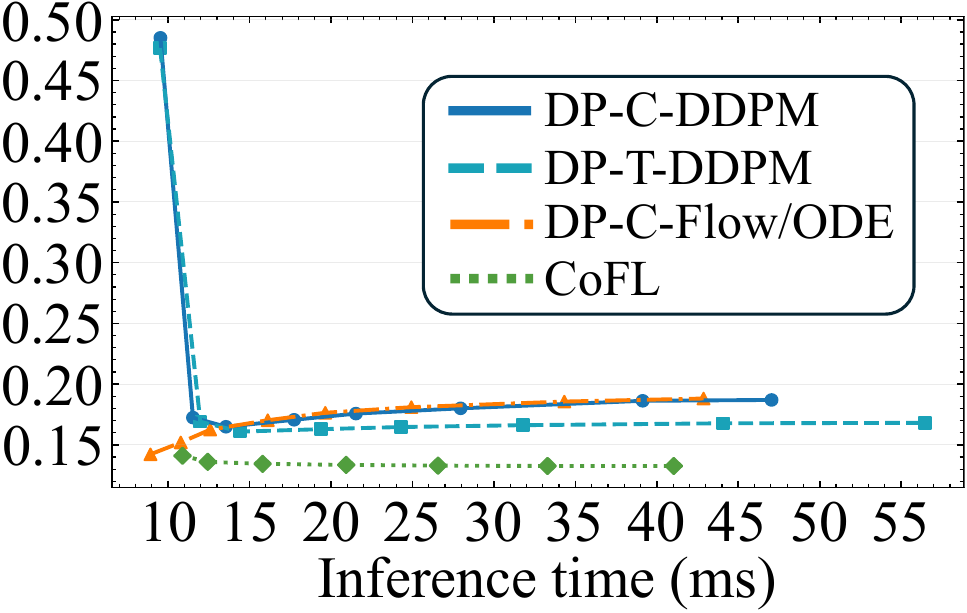}
        \vspace{-5mm}
        \caption{FGE$\downarrow$}
        \label{fig:fge}
    \end{subfigure}
    \hfill
    \begin{subfigure}[b]{0.24\linewidth}
        \centering
        \includegraphics[width=\linewidth]{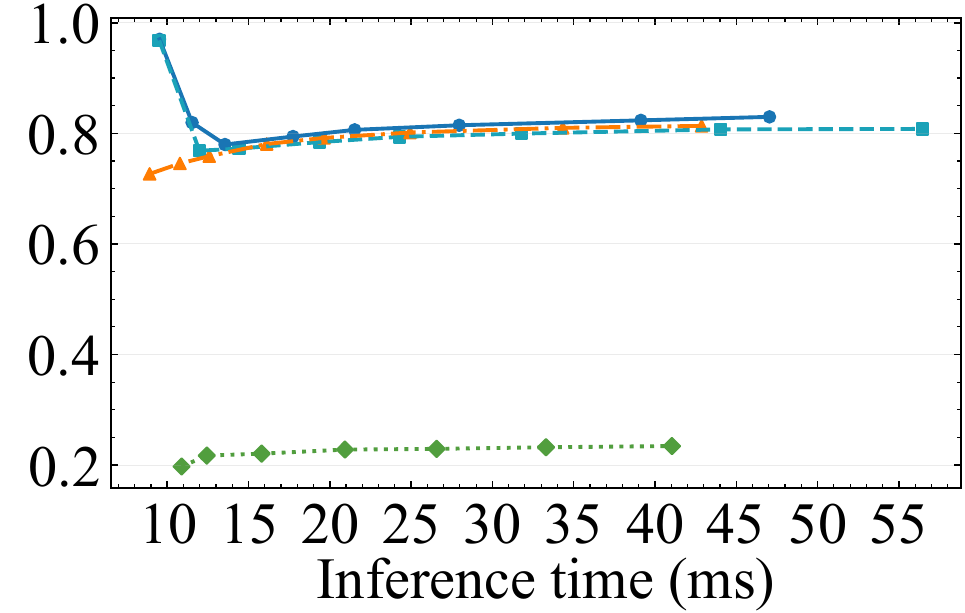}
        \vspace{-5mm}
        \caption{CR$\downarrow$}
        \label{fig:cr}
    \end{subfigure}
    \hfill
    \begin{subfigure}[b]{0.24\linewidth}
        \centering
        \includegraphics[width=\linewidth]{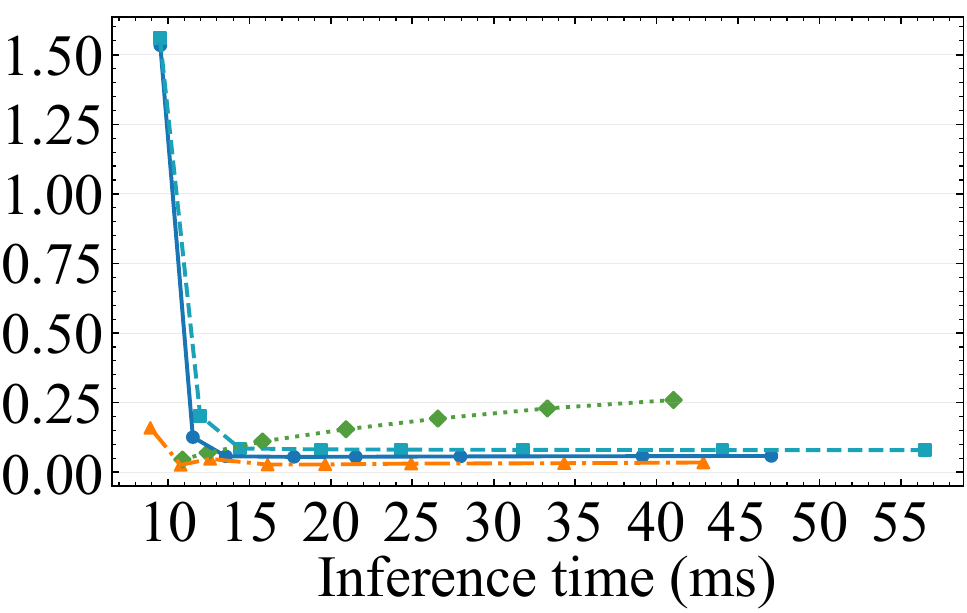}
        \vspace{-5mm}
        \caption{Curv (rad)}
        \label{fig:curv}
    \end{subfigure}
    \hfill
    \begin{subfigure}[b]{0.244\linewidth}
        \centering
        \includegraphics[width=\linewidth]{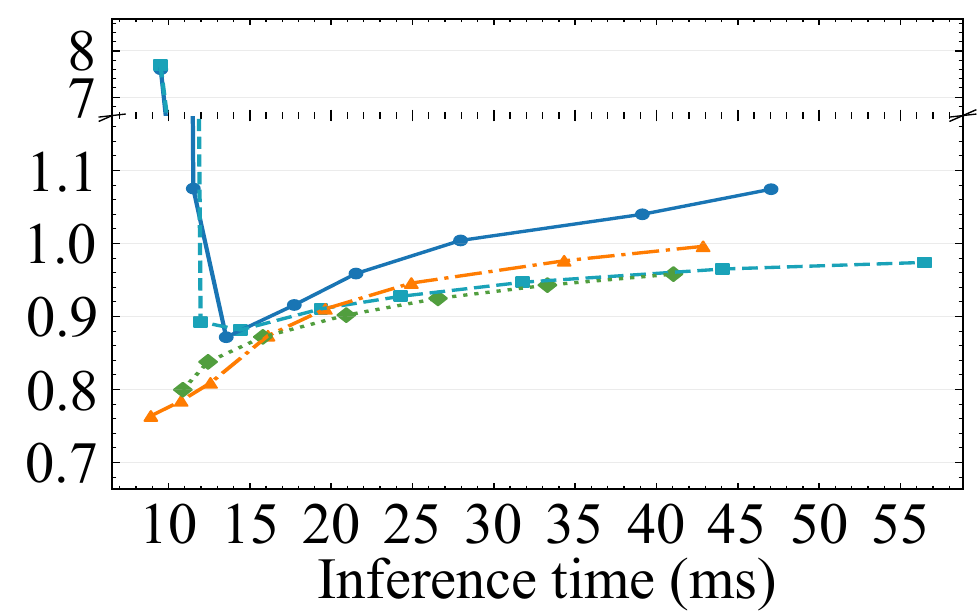}
        \vspace{-5mm}
        \caption{PLR (broken y-axis)}
        \label{fig:plr}
    \end{subfigure}
  \vspace{-2mm}
  \caption{
  {Latency–performance trade-offs of real-time models (DP-family and CoFL).} DP-family: \{1, 2, 3, \underline{5}, 7, 10, 15, 20\} denoising steps (horizon 100); CoFL: \{50, 75, \underline{100}, 125, 150, 175, 200\} grid size (100-step Euler rollout). Underlines mark the operating points reported in Table~\ref{tab:main_results}.
  }
  \label{fig:metrics}
  \vspace{-2mm}
\end{figure*}

\noindent\textbf{Trajectory generative policies (DP family).}
All Diffusion Policy (DP)~\cite{chi2025diffusion} variants output a horizon $T{=}100$ sequence of 2D increments $\Delta\mathbf{x}_{t}$, integrated from $\mathbf{x}_0$ to obtain positions. 
All variants are trained with a fixed 100-step diffusion/flow schedule; at test time we vary the number of sampling steps to control latency (Fig.~\ref{fig:metrics}).
(iii) \emph{DP-C-DDPM}: temporal 1D U-Net with FiLM-style~\cite{perez2018film} global conditioning, sampled with DDPM~\cite{ho2020denoising}. 
We use a 3-stage U-Net (256/512/1024); $\mathbf{x}_0$ is concatenated to the pooled context before FiLM.
(iv) \emph{DP-T-DDPM}: Transformer-based~\cite{ashish2017attn} denoiser with token-level conditioning (no pooling), also sampled with DDPM. 
We use an 8-layer, 4-head Transformer with $d{=}768$; $\mathbf{x}_0$ is projected to $d$ and appended as an extra context token.
(v) \emph{DP-C-Flow/ODE}: same U-Net as (iii), but sampled with a flow/ODE solver~\cite{lipman2023flow, liu2023flow}.
See Appendix~\ref{app:baselines} for implementation details and Fig.~\ref{fig:comparision} for qualitative comparisons.

\begin{table*}[!t]
    \centering
    \caption{{Navigation performance and field quality.} We report trajectory metrics (detailed in \S\ref{subsec:metrics}) aggregated over the validation set for all baselines and CoFL. Field metrics are reported for CoFL only.}
    \vspace{-2mm}
    \label{tab:main_results}
    \resizebox{\linewidth}{!}{
        \begin{tabular}{lccccccccccccccccc}
            \toprule
            \multirow{2}{*}{Method} &
            \multirow{2}{*}{\makecell[c]{Head params (M)}} &
            \multicolumn{4}{c}{Section A\textsuperscript{$\ast$}} &
            \multicolumn{4}{c}{Section B\textsuperscript{$\dagger$}} &
            \multicolumn{4}{c}{Section C\textsuperscript{$\ddagger$}} &
            \multicolumn{2}{c}{Field quality} & \multirow{2}{*}{Lat. (ms)} \\
            \cmidrule(lr){3-6} \cmidrule(lr){7-10} \cmidrule(lr){11-14} \cmidrule(lr){15-16}
             && FGE$\downarrow$ & CR$\downarrow$ & Curv & PLR
             & FGE$\downarrow$ & CR$\downarrow$ & Curv & PLR
             & FGE$\downarrow$ & CR$\downarrow$ & Curv & PLR
             & AE$\downarrow$ & ME$\downarrow$ & \\
            \midrule
            & \multicolumn{14}{c}{\textbf{Matterport3D} (A: 18 regions, 2635 samples; B: 20 regions, 1879 samples; C: 22 regions, 3421 samples)} \\
            \midrule
            Pure VLM & API 
            & 0.22 & 0.90 & 0.03 & 0.84
            & 0.25 & 0.91 & 0.03 & 1.13
            & 0.22 & 0.81 & 0.03 & 1.07
            & --   & -- & --\\
            VLM+Planner & API 
            & 0.24 & 0.84 & 0.06 & 1.05
            & 0.25 & 0.90 & 0.06 & 1.35
            & 0.22 & 0.80 & 0.06 & 1.08
            & --   & -- & --\\
            DP-C-DDPM  &  74.02
            & 0.17 & 0.86 & 0.06 & 0.86
            & {0.19} & 0.79 & 0.05 & 0.94
            & {0.16} & 0.72 & 0.05 & 0.88
            & --   & -- & 17.62\\
            DP-T-DDPM &  81.21
            & \underline{0.17} & 0.86 & 0.08 & 0.84
            & \underline{0.18} & 0.80 & 0.08 & 0.96
            & \underline{0.16} & {0.68} & 0.08 & 0.87
            & --   & -- & 19.28\\
            DP-C-Flow/ODE  &  74.02
            & {0.17} & 0.85 & 0.03 & 0.82
            & {0.19} & 0.78 & 0.03 & 0.89
            & {0.16} & 0.70 & 0.03 & 0.84
            & --   & -- & 15.94\\
            \rowcolor{gray!20}
            CoFL (Ours) &  15.36
            & \textbf{0.15} & \textbf{0.22} & {0.11} & {0.85}
            & \textbf{0.13} & \textbf{0.21} & {0.08} & {0.87}
            & \textbf{0.14} & \textbf{0.17} & {0.14} & {0.87}
            & \textbf{18.65$^\circ$} & \textbf{0.07}  & 15.81 \\
            \midrule
            & \multicolumn{14}{c}{\textbf{ScanNet} (A: 3 regions, 733 samples; B: 3 regions, 132 samples; C: 3 regions, 215 samples)} \\
            \midrule
            Pure VLM & API 
            & 0.24 & 0.96 & 0.04 & 1.30
            & 0.20 & 0.92 & 0.04 & 1.02
            & 0.19 & 0.97 & 0.05 & 1.46
            & --   & -- & --\\
            VLM+Planner & API 
            & 0.25 & 0.94 & 0.06 & 1.38
            & 0.21 & 0.93 & 0.06 & 1.61
            & 0.21 & 0.92 & 0.07 & 1.44
            & --   & -- & --\\
            DP-C-DDPM &  74.02
            & {0.16} & {0.88} & 0.06 & 1.10
            & {0.21} & {0.91} & 0.06 & 1.21
            & {0.19} & {0.87} & 0.06 & 1.14
            & --   & -- & 17.62\\
            DP-T-DDPM &  81.21
            & \underline{0.14} & 0.88 & 0.08 & 1.06
            & \underline{0.19} & 0.95 & 0.09 & 1.25
            & \underline{0.16} & {0.86} & 0.09 & 1.18
            & --   & -- & 19.28\\
            DP-C-Flow/ODE &  74.02
            & {0.16} & 0.88 & 0.03 & 1.05
            & {0.20} & 0.92 & 0.03 & 1.17
            & {0.18} & {0.88} & 0.02 & 1.11
            & --   & -- & 15.94\\
            \rowcolor{gray!20}
            CoFL (Ours) &  15.36
            & \textbf{0.09} & \textbf{0.39} & {0.07} & {0.92}
            & \textbf{0.07} & \textbf{0.35} & {0.06} & {0.91}
            & \textbf{0.09} & \textbf{0.40} & {0.09} & {0.96}
            & \textbf{14.57$^\circ$} & \textbf{0.03} & 15.81 \\
            \bottomrule
        \end{tabular}
    }
    \footnotesize
    \raggedright
    \textsuperscript{$\ast$} Matterport3D Scene: ZMojNkEp431; ScanNet Scene: Scan 0700 (Layouts 00--02). 
    \textsuperscript{$\dagger$} Matterport3D Scene: zsNo4HB9uLZ; ScanNet Scene: Scan 0701 (Layouts 00--02).
    \textsuperscript{$\ddagger$} Matterport3D Scene: Z6MFQCViBuw; ScanNet Scene: Scan 0702 (Layouts 00--02).
    \vspace{-1mm}
\end{table*}

\subsection{Main Results}
\label{subsec:results}
For fairness, trajectories from all methods are resampled to 100 waypoints before computing metrics. All DP family baselines are calculated as the average from 5 different seeds.

\subsubsection{Navigation Performance}
Table~\ref{tab:main_results} reveals a persistent decoupling: trajectory baselines can be near-goal yet unsafe.
Across Matterport3D and ScanNet, they achieve reasonable endpoint error (FGE $\approx$0.14--0.25) but collide in most episodes (CR $\ge$0.68 on Matterport3D; $\ge$0.86 on ScanNet), indicating that waypoint plausibility does not enforce free-space validity along the rollout.

CoFL substantially reduces this gap between endpoint accuracy and safety.
It preserves goal accuracy on Matterport3D (FGE 0.13--0.15) while reducing collisions by $\sim$4$\times$ (CR 0.17--0.22) and maintaining smooth paths (Curv 0.08--0.14).
On ScanNet, it is both more accurate and safer (FGE 0.07--0.09; CR 0.35--0.40), despite using a much smaller head than the trajectory-generation baselines (15M vs.\ 74--81M parameters) and avoiding API-based planning.


\subsubsection{Flow Field Quality and Grid Resolution}
The last two columns of Table~\ref{tab:main_results} measure field fidelity; CoFL attains low AE/ME on both datasets, indicating locally consistent guidance.
Fig.~\ref{fig:metrics} shows that grid resolution mainly controls the effective clearance around obstacles.
With finer grids, the free/obstacle boundary is represented more precisely, so rollouts can follow narrow corridors and track contours more closely, which increases Curv and PLR.
However, CR can slightly rise because the effective safety margin becomes thinner: at higher resolution, the discretized obstacle boundary covers less surrounding area, allowing closer passes.

\subsubsection{Computational Efficiency}
\label{subsec:efficiency}
Fig.~\ref{fig:metrics} reports latency--performance trade-offs ($(I,\ell,\mathbf{x}_0)\rightarrow\tau$). For CoFL, inference time depends on the grid resolution; for DP, it depends on the denoising steps. CoFL is robust across inference budgets: changing grid resolution affects latency but only marginally impacts FGE/CR, indicating stable quality with controllable inference time. 

\subsection{Ablation studies}
\label{subsec:ablation}
We ablate several architectural hyperparameters of CoFL.
All variants are trained and evaluated under the same protocol as the main results. For all ablation experiments, the inference query grid size is set to 100, and the Euler rollout step is set to 100, unless otherwise specified.

\subsubsection{Encoder Fusion Depth}
We vary the fusion depth $L$ of the vision--language encoder stack while keeping the decoder query depth $\tilde{L}=2$ and hidden dimension $d=768$. Encoder parameters are varied and are thus reported here as \textit{Encoder param} (Table~\ref{tab:encode_fusion_depth}).
This ablation isolates how much encoder-side capacity is needed to form instruction-aligned BEV-aligned context.
We report the impact on FGE/CR, AE/ME, and latency.

\begin{table}[ht]
\centering
\caption{Encoder Fusion Depth}
\vspace{-2mm}
\label{tab:encode_fusion_depth}
    \resizebox{\linewidth}{!}{
        \begin{tabular}{lcccccc}
            \toprule
            $L$  & Encoder param (M) & FGE $\downarrow$ & CR $\downarrow$ & AE $\downarrow$ & ME $\downarrow$ & Lat. (ms)\\
            \midrule
            2 & 395.28 & 0.14 & 0.22 & 18.73$^\circ$ & 0.06 & 15.08 \\
            \rowcolor{gray!20}
            4 & 414.18 & 0.13 & 0.22 & 18.16$^\circ$ & 0.06 & 15.81 \\
            6 & 433.08 & 0.13 & 0.24 & 18.76$^\circ$ & 0.06 & 16.21 \\
            \bottomrule
        \end{tabular}
    }
\end{table}

\subsubsection{Decoder Query Depth}
We vary the query depth $\tilde{L}$ of the coordinate-conditioned query decoder while keeping the encoder fusion depth $L=4$ and hidden dimension $d=768$. Decoder parameters are varied and are thus reported here as \textit{Head param}  (Table~\ref{tab:decode_query_depth}).
This controls how much computation is spent per queried location when producing $\mathbf{v}(\mathbf{x}\mid I,\ell)$.
We report the impact on FGE/CR, AE/ME, and latency.

\begin{table}[ht]
\centering
\caption{Decoder Query Depth}
\vspace{-2mm}
\label{tab:decode_query_depth}
    \resizebox{\linewidth}{!}{
        \begin{tabular}{lcccccc}
            \toprule
            $\tilde{L}$  & Head param (M) & FGE $\downarrow$ & CR $\downarrow$ & AE $\downarrow$ & ME$\downarrow$ & Lat. (ms)\\
            \midrule
            1 & 8.27 & 0.14 & 0.23 & 18.55$^\circ$ & 0.06 & 12.41 \\
            \rowcolor{gray!20}
            2 & 15.36 & 0.13 & 0.22 & 18.16$^\circ$ & 0.06 & 15.81 \\
            3 & 22.45 & 0.14 & 0.22 & 18.01$^\circ$ & 0.06 & 20.69 \\
            \bottomrule
        \end{tabular}
    }
\end{table}

\subsubsection{Hidden Dimension}
We vary the model width (hidden dimension) of CoFL while keeping the encoder fusion depth $L=4$ and decoder query depth $\tilde{L}=2$. Overall (encoder and decoder) parameters are varied and are thus reported here as \textit{Overall param}  (Table~\ref{tab:hidden_dimension}).
This changes the representational capacity and parameter count while keeping the architecture/topology fixed.
We report the impact on FGE/CR, AE/ME, and latency.

\begin{table}[ht]
\centering
\caption{Hidden Dimension}
\vspace{-2mm}
\label{tab:hidden_dimension}
    \resizebox{\linewidth}{!}{
        \begin{tabular}{lcccccc}
            \toprule
            $d$  & Overall param (M) & FGE $\downarrow$ & CR $\downarrow$ & AE $\downarrow$ & ME $\downarrow$ & Lat. (ms)\\
            \midrule
            512 & 399.63 & 0.13 & 0.26 & 18.67$^\circ$ & 0.07 & 12.29 \\
            \rowcolor{gray!20}
            768 & 429.54 & 0.13 & 0.22 & 18.16$^\circ$ & 0.06 & 15.81 \\
            1024 & 471.25 & 0.13 & 0.22 & 17.90$^\circ$ & 0.06 & 22.55 \\
            \bottomrule
        \end{tabular}
    }
\end{table}

\subsubsection{Self-Attention in CoFL Decoder}
Our default CoFL decoder disables self-attention among query tokens.
We ablate an augmented variant that enables query self-attention (Table~\ref{tab:self_attention_decoder}). All variants use the same training protocol/data split (encoder depth $L=4$, decoder depth $\tilde{L}=2$, and hidden dimension $d=768$).
We keep the inference query grid fixed at $100\times100$.
Evaluating AE/ME at the original $224\times224$ annotated field resolution would require substantially denser query tokens and become infeasible on a single RTX 4090 when combined with quadratic self-attention memory; therefore, we compute AE/ME on a $100\times100$ evaluation grid obtained by downsampling the $224\times224$ annotated field to ensure a stable and fair comparison across variants.
Thus, AE/ME in Table~\ref{tab:self_attention_decoder} is not directly comparable to Table~\ref{tab:main_results} or other ablation results (only used for within-table comparison).
Decoder parameters are varied and are thus reported as \textit{Head param}.

\begin{table}[ht]
\centering
\caption{Self-Attention in CoFL Decoder}
\vspace{-2mm}
\label{tab:self_attention_decoder}
    \resizebox{\linewidth}{!}{
        \begin{tabular}{lcccccc}
            \toprule
            Variant  & Head param (M) & FGE $\downarrow$ & CR $\downarrow$ & AE $\downarrow$ & ME $\downarrow$ & Lat. (ms)\\
            \midrule
            \rowcolor{gray!20}
            W/O & 15.36 & 0.13 & 0.22 & 17.89$^\circ$ & 0.07 & 15.93 \\
            With & 20.09 & 0.14 & 0.23 & 18.30$^\circ$ & 0.07 & 49.34 \\
            \bottomrule
        \end{tabular}
    }
\end{table}

As shown in Table~\ref{tab:self_attention_decoder}, query self-attention does not improve performance in our setting:
Given this unfavorable efficiency--accuracy trade-off under dense grid querying (self-attention scales as $O(Q^2)$), we omit query self-attention by default.

\subsubsection{Rollout Parameterization}
We ablate the inference-time rule used to integrate the predicted field into a fixed-horizon trajectory. 
All variants use the same trained CoFL model and differ only in how the predicted velocity is normalized during Euler rollout.

\begin{table}[ht]
\centering
\caption{Ablation of rollout parameterization.}
\vspace{-2mm}
\label{tab:rollout_time_param}
    \resizebox{\linewidth}{!}{
        \begin{tabular}{lcccc}
            \toprule
            Variant & FGE $\downarrow$ & CR $\downarrow$ & Curv & PLR\\
            \midrule
            \rowcolor{gray!20}
            Stabilized rescaling ($\frac{\mathbf{v}}{(1-t_k)+\beta\,t_k^{\alpha}}$) & 0.13 & 0.22 & 0.11 & 0.87\\
            No stabilizer ($\frac{\mathbf{v}}{1-t_k}$) & 0.75 & 1.00 & 0.18 & 4.57\\
            Unit-speed rollout ($\mathbf{v}/\|\mathbf{v}\|$) & 0.12 & 0.57 & 1.68 & 3.70\\
            \bottomrule
        \end{tabular}
    }
\end{table}

As shown in Table~\ref{tab:rollout_time_param}, stabilized rescaling gives the best overall trade-off, achieving low goal error and collision rate while preserving smooth and compact trajectories. 
Removing the stabilizer makes the inverse-time scaling numerically unstable near the end of the rollout, causing large terminal steps and frequent collisions. 
In contrast, unit-speed rollout discards the learned magnitude and therefore cannot slow down near the goal; although it attains a slightly lower FGE, it overshoots and oscillates, leading to much higher CR, Curv, and PLR. 
These results indicate that the predicted magnitude is not merely a speed term, but also plays an important role in stable termination and safe fixed-horizon execution.

\subsubsection{Supervision Coverage}
\label{subsec:supervision_coverage}

We ablate the spatial coverage of field supervision to test whether CoFL's gain comes from dense workspace-level learning. 
All variants use the same architecture, dataset, training schedule, and per-step query budget ($N_s=1000$), and differ only in where the training queries are sampled.

\begin{figure}[!t]
  \centering
  \begin{subfigure}[b]{0.32\linewidth}
        \centering
        \includegraphics[width=\linewidth]{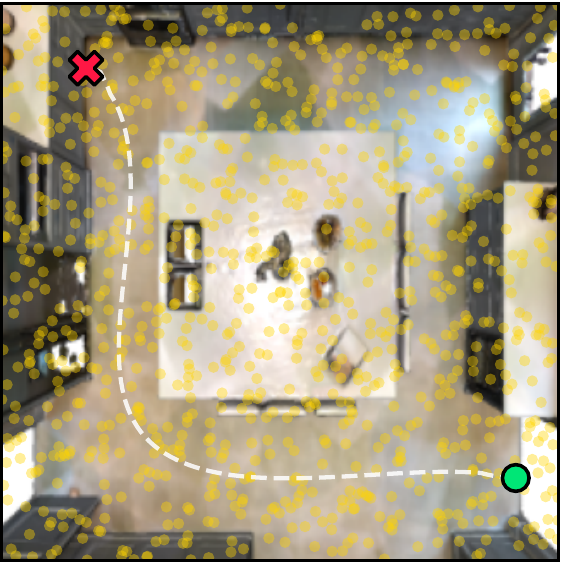}
        \vspace{-5mm}
        \caption{Full workspace}
        \label{fig:full_bev}
    \end{subfigure}
    \hfill
    \begin{subfigure}[b]{0.32\linewidth}
        \centering
        \includegraphics[width=\linewidth]{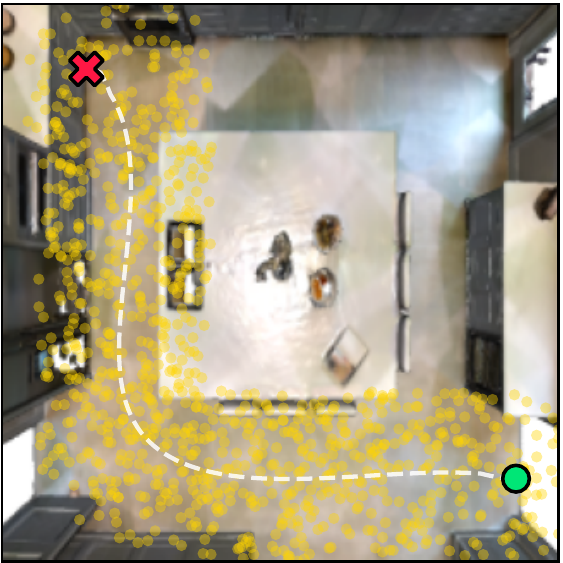}
        \vspace{-5mm}
        \caption{Tube}
        \label{fig:tube}
    \end{subfigure}
    \hfill
    \begin{subfigure}[b]{0.32\linewidth}
        \centering
        \includegraphics[width=\linewidth]{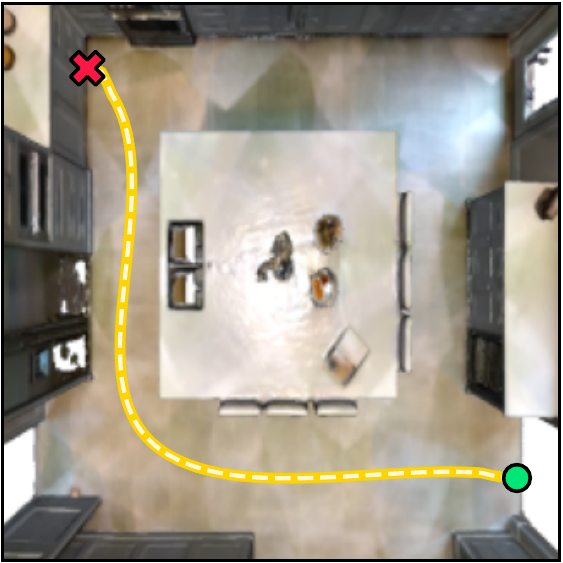}
        \vspace{-5mm}
        \caption{On-trajectory}
        \label{fig:traj}
    \end{subfigure}
  \vspace{-2mm}
  \caption{
  {Example of query distributions (in yellow) for the supervision coverage ablation. {Workspace} supervises the full BEV domain, {Tube} supervises a corridor around the reference path, and {On-trajectory} supervises only the demonstrated trajectory.
  }}
  \label{fig:supervision_sampling}
  \vspace{-1mm}
\end{figure}

We compare three settings (Fig.~\ref{fig:supervision_sampling}). 
(i) \textbf{Full workspace} (Fig.~\ref{fig:full_bev}) is our default strategy, where queries are stratified over the full BEV image. 
(ii) \textbf{On-trajectory} (Fig.~\ref{fig:traj}) samples queries only from the reference trajectory. 
Although this variant still predicts vectors, its effective supervision is restricted to the demonstrated trajectories, reducing data usage to the same trajectory-local regime as the DP-based method. 
(iii) \textbf{Tube} (Fig.~\ref{fig:tube}) is an intermediate setting that samples within a corridor around the reference trajectory, with a half-width of $0.15$ in normalized image coordinates. 
Thus, the three variants compare full 2D workspace supervision, local trajectory-neighborhood supervision, and trajectory-only supervision under the same query budget.

\begin{table}[h]
\centering
\caption{Ablation of supervision coverage. All variants use the same model and query budget, differing only in where training queries are sampled.}
\vspace{-2mm}
\label{tab:supervision_coverage}
\resizebox{\linewidth}{!}{
\begin{tabular}{lcccccc}
\toprule
Variant & FGE $\downarrow$ & CR $\downarrow$ & Curv & PLR & AE $\downarrow$ & ME $\downarrow$\\
\midrule
\rowcolor{gray!20}
Full workspace & 0.13 & 0.22 & 0.11 & 0.87 & 18.73$^\circ$ & 0.06 \\
Tube & 0.12 & 0.61 & 1.66 & 3.70 & 24.14$^\circ$ & 0.08 \\
On-trajectory & 0.34 & 0.91 & 0.52 & 2.48 & 71.71$^\circ$ & 0.46 \\
\midrule
DP-T-DDPM & 0.16 & 0.78 & 0.08 & 0.91 & -- & -- \\
\bottomrule
\end{tabular}
}
\end{table}

\begin{figure*}[!t]
  \centering
  \includegraphics[width=\linewidth]{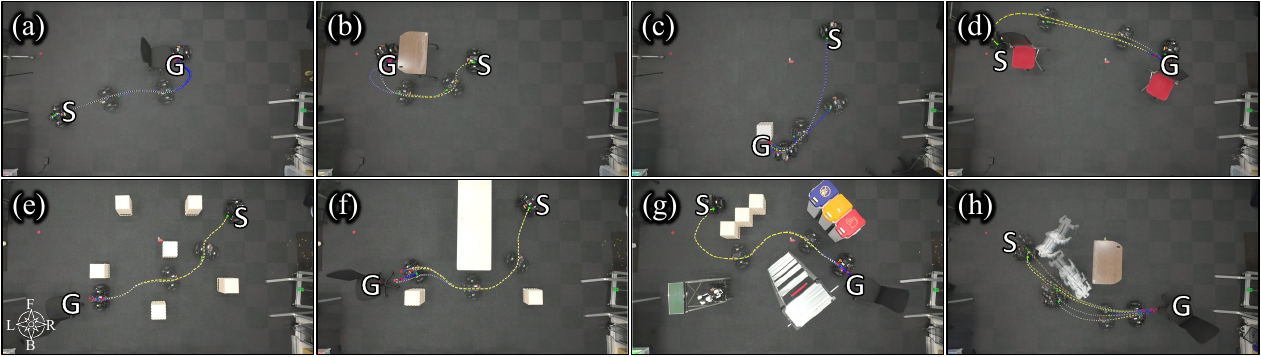}
  \caption{{Example trajectories from real-world experiments}. (a)--(d) \textbf{E1}: from left to right, instructions are ``\textit{move to the right side of the chair}'', ``\textit{move to the left side of the table}'', ``\textit{move to the back side of the box}'', and ``\textit{move to the front side of the right chair}''. (e)--(h) \textbf{E2}: three static layouts and one reactive adversary, all with the same instruction ``\textit{move to the chair}''. Trajectories are replanned at each control step from the latest observation and visualized in the figure.}
  \label{fig:realworld_overview}
  \vspace{-3mm}
\end{figure*}

As shown in Table~\ref{tab:supervision_coverage}, reducing the supervision support substantially degrades navigation quality. 
The tube variant still reaches the goal with low FGE, but its high CR, Curv, and PLR indicate oscillatory and unsafe rollouts, suggesting that supervising only a corridor around the path is insufficient for stable and smooth workspace-level guidance. 
The on-trajectory variant degrades more severely, with much worse FGE, CR, AE, and ME, and even performs worse than DP-T-DDPM (5 denoising steps). 
This shows that simply using a field decoder is not enough: when the supervision collapses to trajectory-local vectors, whose spatial support is comparable to trajectory-based DP-family baselines, CoFL loses its advantage. 
These results confirm that CoFL's main gain comes from converting each scene--instruction pair into dense workspace-level control supervision.

\noindent\textbf{Why does CoFL outperform DP-based trajectory generators?}
Diffusion Policy is a strong generative framework for continuous robot action generation, and our DP baselines can often reach near the target with a smooth rollout, as reflected by competitive FGE and Curv. 
However, their high CR shows that endpoint accuracy alone is insufficient for geometry-sensitive navigation. 
Changing the sampler or denoiser does not close the gap, and the same trend holds under pseudo receding-horizon evaluation (Appendix~\ref{app:pseudo_rh}), suggesting that the issue is not simply sampling or open-loop execution.
CoFL achieves safer trajectories with a much simpler cross-attention decoder because it extracts more supervision signal from each scene-instruction annotation.
DP learns a start-anchored sequence of displacement vectors, whereas CoFL supervises position-conditioned vectors across the BEV-aligned workspace. 
Our supervision coverage ablation (Table~\ref{tab:supervision_coverage}) supports this explanation: when CoFL is reduced to near-trajectory or trajectory-only supervision, performance drops sharply, with trajectory-only supervision performing even worse than DP. 
This confirms that the gain comes from dense workspace-level guidance rather than the decoder alone.

\begin{table*}[!t]
\centering
\caption{{Real-world target navigation under two settings}.
Each setting contains four task cases (C1--C8; 5 trials per case). PN denotes penetration depth in collided trials, while CLR denotes the minimum obstacle clearance in non-collision trials.}
\vspace{-2mm}
\label{tab:realworld_results}
\resizebox{\linewidth}{!}{%
\begin{tabular}{lccccccc}
\toprule
Setting & Case & On-Target$\uparrow$ & CR$\downarrow$ & PN (m) & CLR (m) & TTS (s) & PL (m)  \\
\midrule
\multirow{5}{*}{E1: W/O Obstacles}
& C1: Target = \textit{chair} & 1.00 & --   & --   & --   & 14.13 & 2.83 \\
& C2: Target = \textit{table} & 0.80 & --   & --   & --   & 13.37 & 2.79  \\
& C3: Target = \textit{box}  & 0.60 & --   & --   & --   & 33.88 & 4.79  \\
& C4: Target = \textit{chair pair} (language disambiguation)  & 1.00 & --   & --   & --   & 18.33 & 4.04  \\
& \textbf{Avg} & \textbf{0.85} & \textbf{--} & \textbf{--} & \textbf{--} & \textbf{18.67} & \textbf{3.52}  \\
\midrule
\multirow{5}{*}{E2: With Obstacles}
& C5: Target = \textit{chair}, Static layout A & 1.00 & 0.20 & 0.12 & 0.04 & 22.69 & 3.78 \\
& C6: Target = \textit{chair}, Static layout B & 1.00 & 0.00 & -- & 0.07 & 28.32 & 4.06 \\
& C7: Target = \textit{chair}, Static layout C & 1.00 & 0.20 & 0.03 & 0.05 & 34.39 & 4.90 \\
& C8: Target = \textit{chair}, Reactive adversary & 1.00 & 0.80 & 0.06 & 0.11 & 33.78 & 4.05 \\
& \textbf{Avg} & \textbf{1.00} & \textbf{0.30} & \textbf{0.07} & \textbf{0.06} & \textbf{29.79} & \textbf{4.20} \\
\bottomrule
\end{tabular}%
}
\end{table*}

\begin{figure*}[!t]
  \centering
  \begin{subfigure}[b]{0.24\linewidth}
        \centering
        \includegraphics[width=\linewidth]{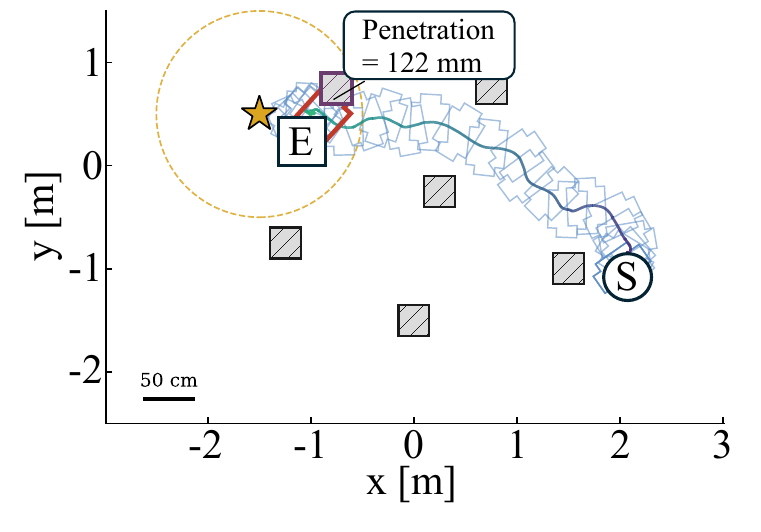}
        \vspace{-5mm}
        \caption{Visualization for failure in C5.}
        \label{fig:failurea}
    \end{subfigure}
    \hfill
    \begin{subfigure}[b]{0.24\linewidth}
        \centering
        \includegraphics[width=\linewidth]{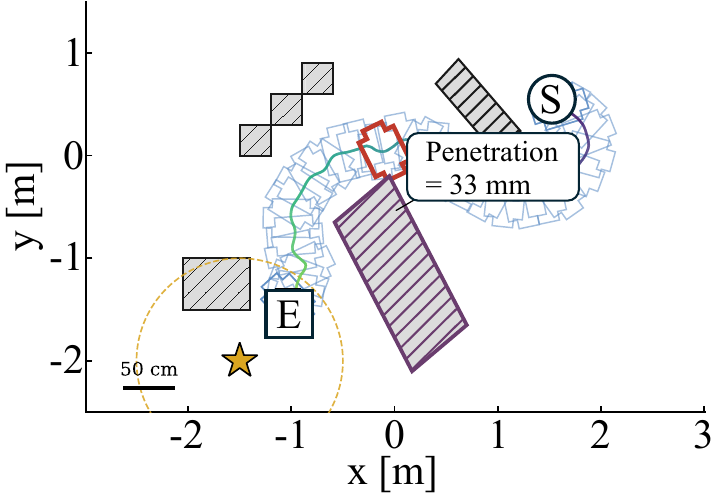}
        \vspace{-5mm}
        \caption{Visualization for failure in C7.}
        \label{fig:failureb}
    \end{subfigure}
    \hfill
    \begin{subfigure}[b]{0.24\linewidth}
        \centering
        \includegraphics[width=\linewidth]{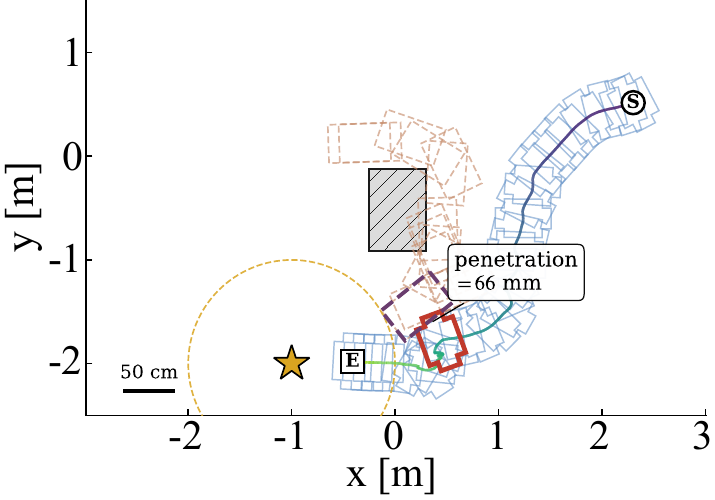}
        \vspace{-5mm}
        \caption{Visualization for failure in C8.}
        \label{fig:failurec}
    \end{subfigure}
    \hfill
    \begin{subfigure}[b]{0.24\linewidth}
        \centering
        \includegraphics[width=\linewidth]{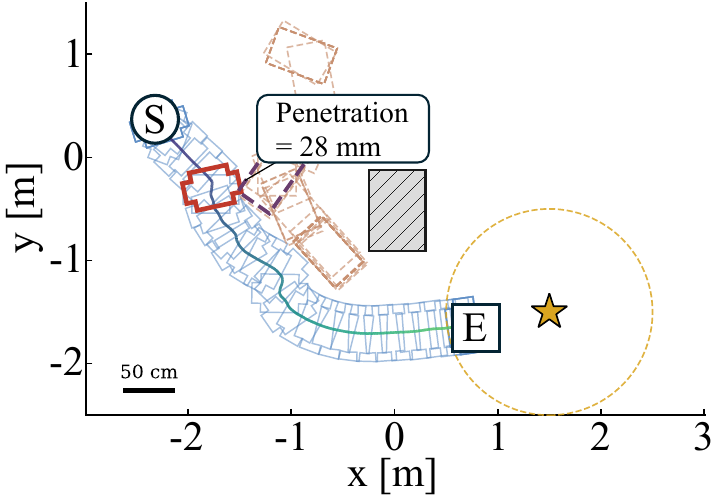}
        \vspace{-5mm}
        \caption{Visualization for failure in C8.}
        \label{fig:failured}
    \end{subfigure}
  \vspace{-1mm}
  \caption{Four collision trials drawn from E2 as examples.
    Trajectories are colored by elapsed time from start to end. 
    The static obstacle is shown with diagonal hatching, and the moving adversary is shown with a dashed orange outline. 
    The deepest-penetration frame is highlighted by the red robot outline, with the penetration depth annotated.}
  \label{fig:failure_example}
  \vspace{-2mm}
\end{figure*}

\section{Zero-Shot Transfer to Real-World Navigation}
\label{sec:realworld}

We deploy CoFL in indoor calibrated top-down observation setups to evaluate its zero-shot transferability in closed-loop semantic navigation.
Notably, we directly use the model trained in \S\ref{sec:experiments} without any real-world fine-tuning.

\subsection{System Setup}
\label{subsec:realworld_setup}
We evaluate in a $6{\times}4$\,m indoor arena monitored by a single overhead camera (DJI OSMO ACTION 4, 30\,Hz, $1280{\times}720$).
At 30\,Hz control loop, CoFL takes $(I,\ell,\mathbf{x}_0)$: a BEV image $I$, a user-typed instruction $\ell$, and the robot position $\mathbf{x}_0$ from motion capture, mapped to image coordinates via a calibrated planar homography.
To reduce self-occlusion, we blur-mask the robot-body bounding box derived from its footprint.
We numerically integrate the predicted flow field to generate a trajectory, which is then tracked by a differential-drive base with a footprint of 0.37 $\times$ 0.54\,m (length $\times$ width). Deployment uses ONNX Runtime on a laptop workstation (Intel Core i9-13900H; RTX 4090 Laptop GPU, 80\,W TGP) with a $100{\times}100$ query grid and a 50-step Euler rollout. Static objects are placed at fixed recorded poses; dynamic obstacles (e.g., robots) are motion-capture tracked; all states share the same world frame.

\subsection{Tasks and Metrics}
\label{subsec:realworld_tasks}
Two navigation settings are evaluated:
\textbf{E1}: without obstacles; comprises 20 trials ({4 target types} $\times$ {5 relative-direction commands}), where each case C1--C4 corresponds to one target type and aggregates the five direction commands (left/right/front/back/NONE). \textbf{E2}: with obstacles; comprises 20 trials ({1 target type} $\times$ {5 trials} $\times$ ({3 static layouts} $+$ {1 reactive adversary})), where cases C5--C7 correspond to the three static obstacle layouts, and C8 uses a reactive adversary (dynamic obstacle).
We report navigation accuracy (\textbf{On-target}: within 60 s, the robot reaches within 1.0 m of the target and remains stationary (speed $<$ 0.01 m/s) for at least 3 s. If a direction is specified, the robot’s final position must also agree with the desired direction (180° acceptance sectors)), safety under E2 (\textbf{CR}: collision rate, where a collision is triggered when the robot's 2D collision footprint intersects an obstacle's 2D collision footprint; both collision footprints are modeled based on their physical sizes; \textbf{PN}: mean penetration depth over collided trials only; \textbf{CLR}: mean minimum obstacle clearance over non-collision trials only), efficiency (\textbf{TTS}: time-to-success, \textbf{PL}: path length), and latency (\textbf{Lat.}: average latency measured from image frame reception to trajectory command publication) (See Fig.~\ref{fig:realworld_overview} for examples). When aggregated over E2, PN and CLR are averaged over their respective subsets of trials rather than over all 20 trials.

\subsection{Results}
\label{subsec:realworld_results}

Table~\ref{tab:realworld_results} summarizes the real-world performance of CoFL.
Across all real-world trials, CoFL runs with low inference latency (28.57\,ms), enabling stable closed-loop replanning.
In the obstacle-free setting (E1), CoFL reaches the target in 85\% of trials, with an average time-to-success of 18.67\,s and a path length of 3.52\,m.
The main failures are concentrated in C3, where the target is a \textit{box}. This case shows the longest TTS among C1--C4 and a substantially longer PL, but the increased motion does not lead to faster completion. The resulting lower effective progress rate (PL/TTS) suggests that the robot spends more time making limited net progress. In practice, we observe frequent heading changes and occasional turn-backs near decision points, which accumulate delay and increase the chance of exceeding the 60\,s time budget. We attribute this behavior to the \textit{box} category being small and harder to disambiguate from the overhead view, as well as being more weakly supervised in the training data. A practical mitigation is to collect a small amount of target-environment data and perform lightweight fine-tuning or adapter-based tuning to improve category-level recognition for small objects.

In the obstacle setting (E2), CoFL achieves a 100\% on-target rate with an overall collision rate of 0.30. 
Despite the nonzero CR, the contact severity remains low: averaged over collided trials, the penetration depth is PN$=0.07$\,m. 
For successful avoidance trials, CoFL maintains a minimum obstacle clearance of CLR$=0.06$\,m on average, indicating close but feasible obstacle negotiation. 
This behavior comes at the cost of longer routes and execution time (TTS: 29.79\,s, PL: 4.20\,m).

The collisions are mainly concentrated in C8, the reactive-adversary case (CR=0.80), whereas the three static layouts remain substantially more reliable (CRs of 0.20, 0.00, and 0.20 for C5--C7). 
In C8, the high CR is accompanied by low penetration depth (PN$=0.06$\,m), suggesting that the robot often reacts to the incoming obstacle and attempts to evade it, but may still experience brief grazing contacts rather than severe collisions. 
Representative examples are provided in Fig.~\ref{fig:failure_example}. 
These failures are partly caused by the nonholonomic differential-drive base, which limits rapid lateral evasive maneuvers when an obstacle reacts at close range. 
In addition, the current perception pipeline introduces extra latency due to image transmission through the camera, edge computer, ROS forwarding, and inference workstation, which can delay short-horizon reactions. 
In future work, running inference directly on the image-receiving edge computer would reduce the end-to-end perception-to-trajectory latency and improve responsiveness in reactive-obstacle scenarios.


\section{Conclusion}
\label{sec:conclusion}

We introduced CoFL, a language-conditioned navigation policy that predicts a continuous flow field over BEV-aligned workspaces and generates trajectories by integrating the field from the robot state.
Rather than learning a start-anchored trajectory per sample, CoFL learns local motion vectors over the workspace, turning each scene--instruction pair into dense spatial control supervision.
This field-based formulation enables more precise and safer motion, along with simple rollout through numerical integration with controllable real-time inference.
Across the large scale benchmark constructed from Matterport3D and ScanNet, CoFL achieves strong scene-wise generalization under strict disjoint validation splits, outperforming modular VLM pipelines and generative trajectory baselines with a lightweight field decoder.
Moreover, the learned policy transfers to real-world setups without fine-tuning, supporting closed-loop navigation across multiple layouts.

\noindent\textbf{Limitations and future work.}
(i) Our real-world validation currently focuses on a controlled calibrated-camera setting, where BEV observations and robot states are externally available; future work may evaluate CoFL with BEV constructed from aerial--ground collaboration and onboard multi-view perception, thereby testing robustness to more dynamic perception noise.
(ii) Current semantics-derived supervision fixes a single clearance profile; future work may enable adjustable safety behavior by training with varied margins and conditioning the model on the desired safety margin at inference.
(iii) CoFL currently operates on 2D BEV-aligned workspaces; extending the field representation to ego-centric sector fields or 3D spatial abstractions may broaden the framework toward first-person navigation and more general vision-language control.



\bibliographystyle{IEEEtran}
\bibliography{citation.bib}

\section*{Supplementary Material}
For completeness, we provide additional appendix materials in the supplementary document. 
The supplementary material includes implementation details of procedural annotation (Appendix~\ref{app:gt_flow}) and benchmark metrics (Appendix~\ref{app:metrics}), extended baseline descriptions (Appendix~\ref{app:baselines}), additional qualitative results (Appendix~\ref{app:more_example_benchmark}), and DP pseudo receding-horizon analysis (Appendix~\ref{app:pseudo_rh}). 

\clearpage

\appendices

\section{Details of Procedural Annotation}
\label{app:gt_flow}

This appendix describes the implementation details of Algorithm~\ref{alg:gt_generation} for deriving the flow field $\mathbf{V}^*$ and the reference trajectory $\tau^*$.
For clarity, we distinguish pixel-grid coordinates $\mathbf{p}\in\{0,\dots,W\!-\!1\}\times\{0,\dots,H\!-\!1\}$ from normalized continuous coordinates $\mathbf{x}\in[0,1]^2$ used in the main text.
Grid-defined maps (masks and fields) are queried at continuous $\mathbf{x}$ via standard bilinear interpolation on the $H{\times}W$ raster.
Distances are computed in pixel units and converted to normalized coordinates when forming $\mathbf{V}^*$ or storing $\tau^*$.

\subsection{Stage 1: Traversability and Goal Sources}
\label{app:gt_stage1}

\subsubsection{Free/Obstacle Masks}
Given the semantic map $\mathcal{S}$, \textbf{\textsc{ExtractFree}} constructs a binary free-space mask $\mathcal{M}_{\mathrm{free}}\in\{0,1\}^{H\times W}$ using the predefined dataset-specific label mapping $\mathcal{R}$.
Pixels whose semantic labels belong to the free set are marked as $\mathcal{M}_{\mathrm{free}}{=}1$, and we define $\mathcal{M}_{\mathrm{obs}}=\neg\mathcal{M}_{\mathrm{free}}$.

\subsubsection{Goal Sources}
\textbf{\textsc{ComputeGoal}} returns goal source pixels $\mathbf{p}_g$ from the target instance specified by $\ell_{\mathrm{target}}$, typically as a thin boundary band adjacent to the target in free space.
When $\ell_{\mathrm{target}}$ contains directional modifiers, such as left/right or front/back, the goal source is restricted to the corresponding side of the target boundary in the BEV coordinate frame.
If no valid goal source exists (e.g., the target is fully isolated from free space), we fall back to the nearest free pixel to the target center as a singleton goal source.

\subsection{Stage 2: Cost-weighted Geodesic Distance}
\label{app:gt_stage2}

\subsubsection{Distance-to-obstacle Transform}
\textbf{\textsc{DTO}} computes the Euclidean distance transform $D_{\mathrm{free}}$ over free space $\mathcal{M}_{\mathrm{free}}=1$, where $D_{\mathrm{free}}(\mathbf{p})$ is the distance (in pixels) from a free pixel $\mathbf{p}$ to the nearest obstacle pixel.

\subsubsection{Safety-aware Cost Map}
\textbf{\textsc{CostMap}} converts $D_{\mathrm{free}}$ into a traversal cost map $C_{\mathrm{cost}}\ge 1$ by applying a truncated linear penalty within a safety band of radius $\rho_{\mathrm{safe}}$:
\begin{equation}
C_{\mathrm{cost}}(\mathbf{p})
=
1 + \lambda_{\mathrm{safe}}\,[\rho_{\mathrm{safe}} - D_{\mathrm{free}}(\mathbf{p})]_+,
\label{eq:costmap}
\end{equation}
where $[z]_+=\max(0,z)$. Intuitively, this increases costs near obstacles and encourages paths with larger clearance.

\subsubsection{Cost-weighted Geodesic and Predecessor Map}
\textbf{\textsc{Geodesic}} runs the Dijkstra~\cite{dljkstra1959note} on an 8-connected pixel grid, restricted to free cells ($\mathcal{M}_{\mathrm{free}}{=}1$).
We treat each free pixel as a graph node.
For any two neighboring pixels $\mathbf{p}$ and $\mathbf{q}$ (axial/diagonal neighbors), we assign an edge cost
\begin{equation}
w(\mathbf{p},\mathbf{q})
=
\tfrac12\bigl(C_{\mathrm{cost}}(\mathbf{p})+C_{\mathrm{cost}}(\mathbf{q})\bigr) \|\mathbf{p}-\mathbf{q}\|_2,
\end{equation}
where $\|\mathbf{p}-\mathbf{q}\|_2\in\{1,\sqrt{2}\}$ for axial/diagonal moves.
Starting from all goal pixels as sources, Dijkstra~\cite{dljkstra1959note} produces (i) a cost-weighted distance-to-go field $D_g^w(\mathbf{p})$ and (ii) a predecessor pointer $\mathrm{pred}(\mathbf{p})$, which stores the next pixel on the lowest-cost route from $\mathbf{p}$ to the goal.
Collectively, $\mathrm{pred}(\cdot)$ forms a shortest-path tree rooted at the goal sources.

\subsubsection{Pixel Distance-to-go Along the Predecessor Tree}
\textbf{\textsc{PixelLengthFromPred}} computes $D_g^{\mathrm{pix}}(\mathbf{p})$, defined as the geometric remaining path length (in pixels) when repeatedly following the predecessor pointers from $\mathbf{p}$ to a goal.
Equivalently, it is the accumulated step length along the predecessor chain:
\begin{align}
D_g^{\mathrm{pix}}(\mathbf{p})
&=
\sum_{k=0}^{K(\mathbf{p})-1}
\bigl\|\mathbf{p}_k-\mathbf{p}_{k+1}\bigr\|_2,\notag\\
\mathbf{p}_0&=\mathbf{p},\ 
\mathbf{p}_{k+1}=\mathrm{pred}(\mathbf{p}_k),
\end{align}
where $\mathbf{p}_{K(\mathbf{p})}$ is the first goal pixel reached.
In implementation, we compute it via the recursion
$
D_g^{\mathrm{pix}}(\mathbf{p})=\|\mathbf{p}-\mathrm{pred}(\mathbf{p})\|_2 + D_g^{\mathrm{pix}}(\mathrm{pred}(\mathbf{p}))
$
(with $D_g^{\mathrm{pix}}(\mathbf{p})=0$ for goal pixels), processing pixels in nondecreasing order of $D_g^w$.

\subsection{Stage 3: Obstacle Inner Repulsion}
\label{app:gt_stage3}

\textbf{\textsc{DTF}} computes an obstacle-space distance transform $D_{\mathrm{obs}}$ that is positive only inside obstacles ($\mathcal{M}_{\mathrm{obs}}{=}1$), where $D_{\mathrm{obs}}(\mathbf{x})$ measures the distance (in pixels) from an obstacle pixel $\mathbf{p}$ to the nearest free pixel.
This term is used solely to produce outward gradients for pixels that lie inside obstacles.

\subsection{Stage 4--5: Potential Field and Flow Field}
\label{app:gt_stage45}

\subsubsection{Potential Construction}
We form a piecewise potential $\Phi$ following Algorithm~\ref{alg:gt_generation}:
\begin{equation}
\Phi(\mathbf{p}) =
\begin{cases}
w_g \, D_g^w(\mathbf{p}), & \mathcal{M}_{\mathrm{free}}(\mathbf{p})=1,\\[2pt]
w_{\mathrm{obs}}\,D_{\mathrm{obs}}(\mathbf{p}) + b_{\mathrm{obs}}, & \mathcal{M}_{\mathrm{obs}}(\mathbf{p})=1.
\end{cases}
\label{eq:phi_piecewise}
\end{equation}
We set $w_{\mathrm{obs}} \gg w_g$ and choose $b_{\mathrm{obs}}=\max_{\mathbf{p}\,:\,\mathcal{M}_{\mathrm{free}}(\mathbf{p})=1} w_gD_g^w(\mathbf{p})$ such that the obstacle-side potentials dominate around the interface, avoiding discrete gradients that point into obstacles.

\subsubsection{Direction and Magnitude}
We smooth $\Phi$ with a Gaussian filter and compute spatial derivatives using Sobel operators.
The unit direction field is
\begin{equation}
\mathbf{u}(\mathbf{p})=\frac{-\nabla \Phi(\mathbf{p})}{\|\nabla \Phi(\mathbf{p})\|_2+\epsilon}.
\end{equation}

In free space, we scale the magnitude by the pixel distance-to-go and convert it to normalized coordinates:
$\mathbf{V}^*(\mathbf{p}) =
\bigl[u_x(\mathbf{p})\cdot D_g^{\mathrm{pix}}(\mathbf{p})/W,\ 
u_y(\mathbf{p})\cdot D_g^{\mathrm{pix}}(\mathbf{p})/H\bigr],$ for $\mathcal{M}_{\mathrm{free}}(\mathbf{p})=1$.
Inside obstacles, we use unit-speed escape along the same direction: $\mathbf{V}^*(\mathbf{p})=\mathbf{u}(\mathbf{p})$ for $\mathcal{M}_{\mathrm{obs}}(\mathbf{p})=1$.

\subsection{Stage 6: Reference Trajectory Extraction}
\label{app:gt_stage6}

\subsubsection{Reachable Free Space and Start Sampling}
Although $\mathcal{M}_{\mathrm{free}}$ marks all non-obstacle pixels, some free regions may be disconnected from the goal sources (e.g., being fully enclosed by obstacles).
Therefore, \textbf{\textsc{SampleStart}} samples the start pixel $\mathbf{p}_0$ only from the reachable subset, defined by a finite distance-to-go:
$D_g^{\mathrm{pix}}(\mathbf{p}_0)<\infty$.
The sampled start is further required to be sufficiently far from the goal sources and obstacles, according to the dataset generation configuration.

\subsubsection{Backtracking and Resampling}
Given $\mathbf{p}_0$, \textbf{\textsc{BacktrackPred}} backtracks predecessors $\mathbf{p}_{k+1}=\mathrm{pred}(\mathbf{p}_k)$ until reaching a goal source to obtain a polyline $\tau_{\mathrm{raw}}$.
\textbf{\textsc{Resample}} then resamples $\tau_{\mathrm{raw}}$ by arc length to a fixed number of waypoints to obtain $\tau^*$, which is stored in normalized coordinates $(p_x/W,\ p_y/H)$.

\section{Details of Evaluation Metrics and Protocol}
\label{app:metrics}
We describe the evaluation protocol and metric implementations used in \S\ref{sec:experiments}.
A trajectory is a sequence of 2D normalized positions $\tau=\{\mathbf{x}_i\}_{i=0}^{N-1}$, where $\mathbf{x}_i=(u_i,v_i)\in[0,1]^2$.
Collision checks use a binary obstacle mask $\mathcal{M}_{\mathrm{obs}}\in\{0,1\}^{H\times W}$, where $\mathcal{M}_{\mathrm{obs}}[y,x]=1$ indicates an obstacle cell.
Given a normalized point $\mathbf{x}=(u,v)$, we first clamp it to $[0,1]^2$ and map it to integer grid indices
\begin{align}
p_x=&\mathrm{clip}(\lfloor uW\rfloor,0,W\!-\!1),\notag\\
p_y=&\mathrm{clip}(\lfloor vH\rfloor,0,H\!-\!1),
\end{align}
then query occupancy using row-major indexing $\mathcal{M}_{\mathrm{obs}}[p_y,p_x]$.

\subsection{Trajectory Resampling}
To make metrics comparable across methods with different waypoint counts, trajectories are resampled to a fixed number of points $K=100$ using piecewise-linear interpolation at uniform arc-length.
We denote the resampled predicted trajectory by $\bar\tau=\{\mathbf{\bar x}_j\}_{j=0}^{K-1}$.
Concretely, let the segment lengths be $d_i=\|\mathbf{x}_{i}-\mathbf{x}_{i-1}\|_2$ for $i=1,\dots,N-1$,
and let the cumulative arc-lengths be $s_0=0$ and $s_i=\sum_{k=1}^{i} d_k$.
We sample $K$ target arc-lengths uniformly in $[0,s_{N-1}]$ and linearly interpolate within the corresponding segment.

\subsection{Trajectory Metrics}
Trajectories are evaluated using common trajectory metrics as follows:

\subsubsection{Final Goal Error (\textbf{FGE})}
Let $\mathbf{x}_\text{end}\in[0,1]^2$ be the endpoint of the annotated trajectory $\tau^*$.
FGE is the Euclidean distance between the final resampled point and the endpoint:
\begin{equation}
\mathrm{FGE}(\bar\tau)=\left\lVert \mathbf{\bar x}_{K-1}-\mathbf{x}_\text{end}\right\rVert_2.
\label{eq:fge}
\end{equation}

\subsubsection{Collision Rate (\textbf{CR})}
CR is a binary indicator of whether the resampled predicted trajectory ever enters an obstacle cell:
\begin{equation}
\mathrm{CR}(\bar\tau)
=
\mathbb{I}\Bigl[\exists\, j:\ \mathcal{M}_{\mathrm{obs}}[p_y(\mathbf{\bar x}_j),p_x(\mathbf{\bar x}_j)]=1\Bigr].
\label{eq:cr}
\end{equation}

The benchmark reports the mean of $\mathrm{CR}(\tau)$ across episodes.

\subsubsection{Curvature-based Smoothness (\textbf{Curv})}
Curvature is the mean absolute change in heading angle between consecutive segments of the resampled predicted trajectory.
Let segment vectors be $\Delta_j=\mathbf{\bar x_{j+1}}-\mathbf{\bar x_j}$.
We discard degenerate segments with $\|\Delta_j\|_2\le\epsilon$ and compute headings
\begin{equation}
\psi_j=\mathrm{atan2}(\Delta_{j}^{(v)},\Delta_{j}^{(u)}).
\end{equation}
Curv is then
\begin{equation}
\mathrm{Curv}(\bar\tau)=\frac{1}{M-1}\sum_{j=0}^{M-2}\left|\mathrm{WrapToPi}\!\left(\psi_{j+1}-\psi_j\right)\right|,
\label{eq:curv}
\end{equation}
where $M$ is the number of valid (non-degenerate) segments.

\subsubsection{Path Length Ratio (\textbf{PLR})}
Let the path length of a resampled trajectory be
\begin{equation}
L(\bar\tau)=\sum_{j=0}^{K-2}\left\lVert \mathbf{\bar x_{j+1}}-\mathbf{\bar x_j}\right\rVert_2.
\end{equation}

PLR is defined as the ratio between predicted and annotated trajectory lengths:
\begin{equation}
\mathrm{PLR}(\bar\tau)=\frac{L(\bar\tau)}{L(\tau^*)}.
\label{eq:plr}
\end{equation}

\subsection{Flow Field Metrics}
Flow field metrics are evaluated on the exact grid of the annotated flow.
Let the annotated flow have a spatial resolution $R\times R$ (e.g., $224\times224$ in our real-world setup).
We flatten all grid locations into $N=R^2$ points and reshape both predicted and annotated flows into
$\mathbf{\hat V}, \mathbf{V}^*\in\mathbb{R}^{N\times 2}$, where $\mathbf{\hat V_n}, \mathbf{V_n}^*\in\mathbb{R}^2$ denotes the predicted/annotated vectors at the $n$-th annotated grid cell.
All field metrics below are computed over the full grid.

\subsubsection{Angular Error (\textbf{AE})}
We compute the clipped cosine similarity
\begin{equation}
c_n=\mathrm{clip}\!\left(
\frac{\mathbf{\hat V}_n}{\|\mathbf{\hat V}_n\|_2+\epsilon}\cdot
\frac{{\mathbf{V}_n}^*}{\|{\mathbf{V}_n}^*\|_2+\epsilon},
-1,1\right),
\end{equation}
and define the per-point angular error as $\Delta\phi_n=\arccos(c_n)\cdot\frac{180}{\pi}$ (degrees).
AE is the mean of $\{\Delta\phi_n\}$ over all evaluated points.

\subsubsection{Magnitude Error (\textbf{ME})}
Magnitude error is
\begin{equation}
\mathrm{ME}=\frac{1}{N}\sum_{n=1}^{N}\left|\|\mathbf{\hat V}_n\|_2-\|{\mathbf{V}_n}^*\|_2\right|.
\end{equation}

\section{Details of Baseline Implementations}
\label{app:baselines}

This appendix describes baseline formulations and implementation details as a supplement for \S\ref{subsec:setup}.
All baselines operate on the same BEV observation $I$, language instruction $\ell$, and start position $x_0$.
Learned baselines use the same frozen vision–language encoder as CoFL (\S\ref{sec:method}),
and differ only in the prediction head and the trajectory generation procedure.

\subsection{Pure VLM}
\label{app:baseline_pure_vlm}
This baseline uses a commercial VLM to directly output a waypoint sequence from an RGB BEV image, a language instruction, and the starting coordinate.
No goal location, obstacle mask, or planner is provided.
The VLM must infer a plausible target from the instruction and propose a collision-free trajectory in one shot.

\subsubsection{Prompt}
We use the following system prompt and request a strict JSON response:
\begin{tcolorbox}[promptbox,title={System prompt for Pure VLM baseline}]
\scriptsize\ttfamily
You are a robot navigation policy operating on a TOP-DOWN VIEW (bird's eye view) and a natural-language instruction.

INPUTS

- Image: a TOP-DOWN VIEW (bird's eye view) (RGB-only; no explicit obstacle mask).

- Text: (1) Instruction and (2) START coordinate in normalized image coordinates.

START (AUTHORITATIVE)

- The robot START is provided in text.

- The image may also show a green dot, but if there is any ambiguity, trust the text START.

COORDINATE SYSTEM (NORMALIZED)

- (0.00, 0.00) is the TOP-LEFT corner of the image

- (1.00, 1.00) is the BOTTOM-RIGHT corner

- x increases left -> right; y increases top -> bottom

- All output coordinates MUST have exactly 2 digits after the decimal (e.g., 0.37).

GOAL

Return a smooth 2D trajectory from START to a TARGET consistent with the instruction, while avoiding obstacles inferred from the RGB image.

MAP INTERPRETATION (RGB-ONLY)

- There is no color-coded traversability. You MUST infer traversable vs obstacle regions from visual cues (e.g., occupied structures, walls, furniture-like shapes, cluttered regions, boundaries of open space).

- When uncertain, be conservative: route through visually open, continuous regions and keep margins from occupied structures.

INSTRUCTION GROUNDING

- Use the instruction to choose a target location on the map.

- If the instruction mentions an object category (chair/table/etc.), attempt to locate a plausible instance from the RGB map using shape/position/context cues.

- If multiple candidates exist, choose the one you judge most consistent with the instruction and overall scene layout.

TARGET REQUIREMENTS (NO FALLBACK; MUST ANSWER)

- You MUST always output a target with a valid (x,y) in [0.00, 1.00] x [0.00, 1.00].

- If you are uncertain about the exact target, output your BEST GUESS anyway.

- Indicate uncertainty in "notes" using short ASCII words (e.g., "best guess").

TRAJECTORY REQUIREMENTS (STRICT)

- Output exactly \{N\} waypoints.

- Waypoint \#1 must be at START (exact match preferred; otherwise within 0.02 L2 distance).

- Waypoint \#\{N\} must be at TARGET (exact match preferred; otherwise within 0.02 L2 distance).

- All waypoints must satisfy: 0.00 <= x <= 1.00 and 0.00 <= y <= 1.00.

- Avoid obstacles inferred from the RGB image (do not place waypoints on occupied structures).

- Prefer smooth paths with gentle curvature; avoid zig-zags.

WAYPOINT SPACING (SOFT)

- Prefer roughly even spacing when feasible. Collision avoidance has higher priority.

OUTPUT FORMAT (STRICT JSON ONLY)

Return ONLY a valid JSON object with double quotes and no trailing commas. No extra text.

Schema:

\{\{
  "target": \{\{"name": "<string>", "x": <float>, "y": <float>\}\},
  "trajectory": [
    \{\{"x": <float>, "y": <float>\}\},
    ...
  ],
  "notes": "<max 25 words; ASCII only; no line breaks>"
\}\}

NOTES (STRICT)

- Use only letters/numbers/spaces in notes (no punctuation).

- If target is uncertain, include "best guess".

- If obstacle inference is uncertain, include "conservative".

- Front should be mapped to top and back should be mapped to  bottom
\end{tcolorbox}

\subsubsection{Query Format}
To reduce ambiguity in the start location, we additionally draw a green dot on the input map, while still providing the authoritative start coordinate in text.
The VLM is instructed to trust the text start if the visualization is unclear.

\subsubsection{Decoding and API Settings}
We use Gemini-2.5-Flash with deterministic decoding (temperature $=0$, top-p$=1.0$) and enforce a JSON-only response format.
We set the maximum output budget to 8192 tokens to reduce truncation.
If the output is malformed, we retry up to two times.

\subsubsection{Output Parsing and Waypoint Normalization}
We parse the returned JSON object and extract \texttt{target} and \texttt{trajectory}.
All coordinates are clamped to $[0,1]$.
If the returned trajectory contains fewer than the required number of waypoints, we interpolate along its arclength to obtain exactly $N$ waypoints; if it contains more, we subsample uniformly by index.
This post-processing only standardizes the waypoint count and does not enforce feasibility (e.g., no collision repair).
The final evaluation uses the same resampling protocol as all methods (100 waypoints) as described in \S\ref{subsec:results}.

\subsection{VLM+Planner}
\label{app:baseline_vlm_planner}
This baseline decomposes navigation into (i) scene understanding and (ii) geometric planning.
Given a top-down BEV image and an instruction, a commercial VLM first predicts the target object and obstacle bounding boxes in normalized image coordinates; an A* planner then computes a collision-free path on a discretized occupancy grid.

\subsubsection{VLM Prediction (Target and Obstacles)}
The VLM outputs a JSON object containing the target object (center and bounding box), a list of obstacle objects (each with a center and bounding box), and an optional start estimate.
We use the following system instruction and request JSON-only output:
\begin{tcolorbox}[promptbox,title={System prompt for VLM in VLM+Planner baseline}]
\scriptsize\ttfamily
You are analyzing a TOP-DOWN VIEW (bird's eye view) of an indoor environment for robot navigation.

INPUTS

- Image: a TOP-DOWN VIEW (bird's eye view) (RGB-only; no explicit obstacle mask).

- Text: (1) Instruction.

COORDINATE SYSTEM (NORMALIZED)

- (0.00, 0.00) is the TOP-LEFT corner of the image

- (1.00, 1.00) is the BOTTOM-RIGHT corner

- x increases left -> right; y increases top -> bottom

- All output coordinates MUST have exactly 2 digits after the decimal (e.g., 0.37).

GOAL

Given a navigation instruction, identify:

1. The TARGET OBJECT (the object the instruction refers to) with its center, bounding box, and direction descriptor (left, right, top, bottom, none).

2. Other obstacle objects that might block a path, with their centers and bounding boxes.

3. If multiple candidates exist, choose the one you judge to be most consistent with the instruction and overall scene layout.

Output ONLY valid JSON. Required keys are exactly:
\{\{
    "target": \{\{
        "name": "object\_{name}",
        "center": [0.50, 0.70],
        "bbox": [0.40, 0.60, 0.60, 0.85],
        "direction": "left/right/top/bottom/none",
        "confidence": "high/medium/low"
    \}\},
    "obstacles": [
        \{\{"name": "obstacle1", "center": [0.30, 0.40], 
        "bbox": [0.25, 0.32, 0.38, 0.48]\}\},
        ...
    ],
    "notes": "<max 25 words; ASCII only; no line breaks>"
\}\}

NOTES (STRICT)

- All coordinates must be between 0.0 and 1.0

- bbox format is [xmin, ymin, xmax, ymax] in normalized coords

- Ensure xmin < xmax and ymin < ymax

- If uncertain, still provide your best guess and set confidence to "low"

- Front should be mapped to top and back should be mapped to  bottom
\end{tcolorbox}

\subsubsection{Query Format and Side-of-object Handling}
The user query provides the instruction and requests (i) a target and (ii) obstacles.
If the instruction specifies approaching a side of an object (e.g., \textit{left of} / \textit{right of} / \textit{above} / \textit{below}), we treat the target object's bounding box as an additional forbidden region and set the navigation goal to a point offset from the corresponding side of the target box.
Concretely, given target bbox $\mathbf{b}=[x_{\min},y_{\min},x_{\max},y_{\max}]$ and center $(x_c,y_c)$, we compute
\begin{equation}
(x_g,y_g)=
\begin{cases}
(x_{\min}-\delta,\,y_c) & \text{left},\\
(x_{\max}+\delta,\,y_c) & \text{right},\\
(x_c,\,y_{\min}-\delta) & \text{top},\\
(x_c,\,y_{\max}+\delta) & \text{bottom},\\
(x_c,\,y_c) & \text{none},
\end{cases}
\qquad \delta=0.02,
\end{equation}
followed by clamping to $[0,1]^2$.

\subsubsection{Decoding and API Settings}
We use Gemini-2.5-Flash with deterministic decoding (temperature $=0$, top-p$=1.0$) and a maximum output budget of 8192 tokens.
If the returned JSON is invalid or missing a target bbox/center, we retry up to two times with an explicit JSON-only reminder.

\subsubsection{Geometric Planning (A*)}
We rasterize the predicted obstacle bounding boxes into a binary occupancy grid of size $G\times G$ (default $G=128$) and run A* (8-neighborhood) from the provided start $\mathbf{x}_0$ to the derived goal $(x_g,y_g)$.
To enforce a safety margin, we inflate each obstacle by a fixed pixel radius $r$ (10\,px in the bbox image space), converted to a normalized margin and then to grid cells before rasterization.
Start/goal are required to lie in free space; otherwise they are snapped to the nearest free cell.
The resulting grid path is converted back to normalized coordinates and resampled by arclength to obtain a fixed-length waypoint sequence for evaluation.

\subsection{Diffusion Policy Family}
\label{app:baseline_dp}

We adopt the official Diffusion Policy (DP) codebase~\cite{chi2025diffusion} as representative trajectory generative policies.
All DP-family baselines share the same vision--language encoder as CoFL and differ only in the denoiser (trajectory decoder) architecture and the inference-time sampler (Table~\ref{tab:dp_family}).

\begin{table*}[!htpb]
\centering
\caption{\textbf{DP-family baselines used in this work}. All variants reuse the same SigLIP-based vision--language encoder and differ in the denoiser architecture and sampler.}
\vspace{-1mm}
\label{tab:dp_family}
\resizebox{\linewidth}{!}{%
\begin{tabular}{lccc}
\toprule
Method & Encoder & Trajectory denoiser & Sampler \\
\midrule
DP-C-DDPM & Vision--Language Encoder same as CoFL & Temporal U-Net (1D Conv; FiLM) & DDPM (stochastic) \\
DP-T-DDPM & Vision--Language Encoder same as CoFL & Transformer (token-conditioning) & DDPM (stochastic) \\
DP-C-Flow/ODE & Vision--Language Encoder same as CoFL & Temporal U-Net (1D Conv; FiLM) & ODE (deterministic) \\
\bottomrule
\end{tabular}%
}
\vspace{-1mm}
\end{table*}

\begin{figure*}[!t]
  \centering
  \includegraphics[width=0.8\linewidth]{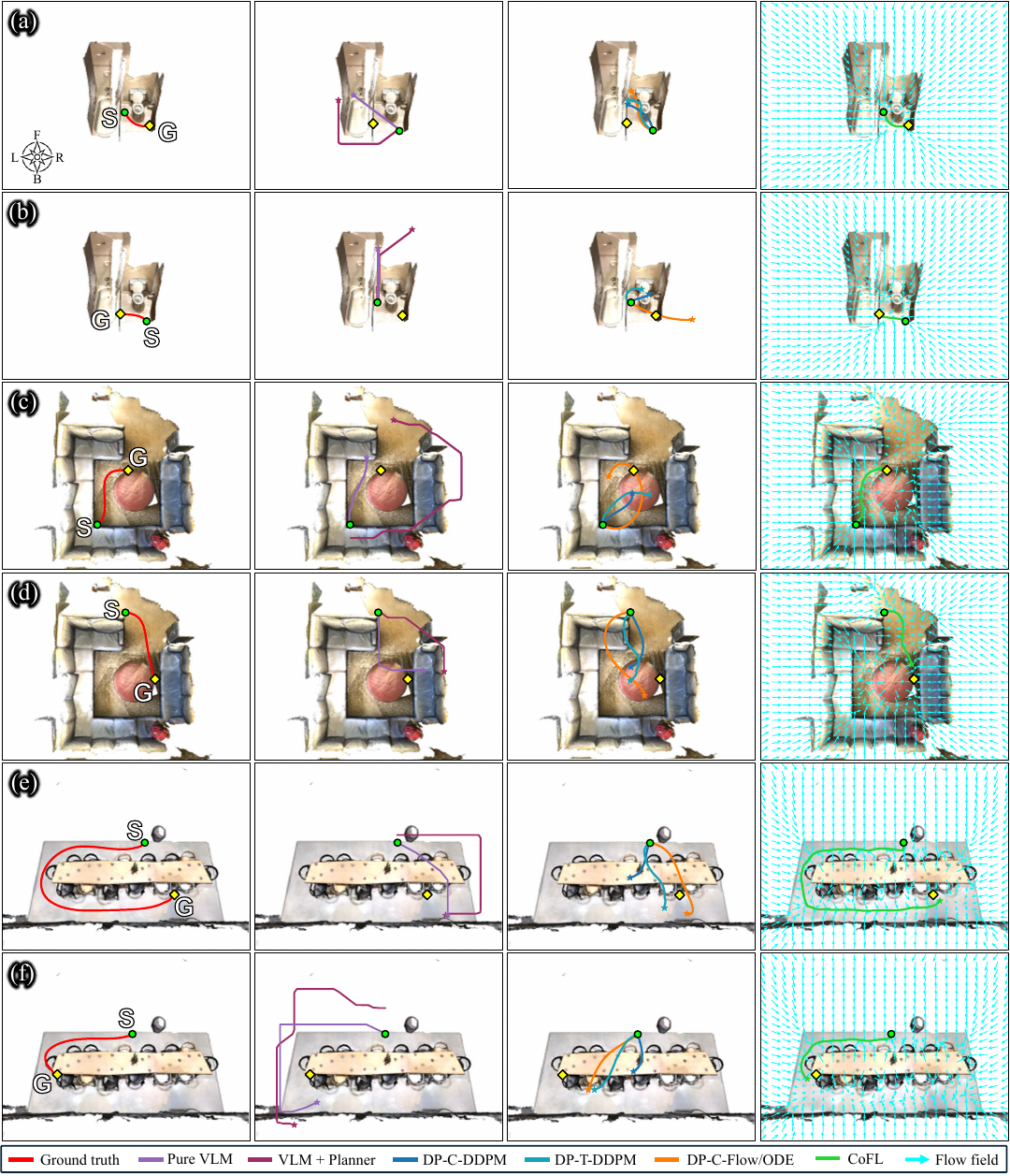}
  \caption{
  \textbf{More examples of trajectories on validation set.}
  From left to right: ground truth, pure VLM/VLM+Planner, DP-family and CoFL. (a-b) ScanNet: \textit{``please walk to the door"} and \textit{``can you approach the bathtub"}; (c-d) ScanNet: \textit{``go to ahead of the table"} and \textit{``can you approach right side of the table"}; (e-f) Matterport3D: \textit{``can you proceed to back of the chair in the lower right"} and \textit{``please travel to the chair in the lower left"}.
  }
  \vspace{-3mm}
  \label{fig:example_benchmark}
\end{figure*}

\begin{figure*}[!t]
  \centering
  \begin{subfigure}[b]{0.22\linewidth}
        \centering
        \includegraphics[width=\linewidth]{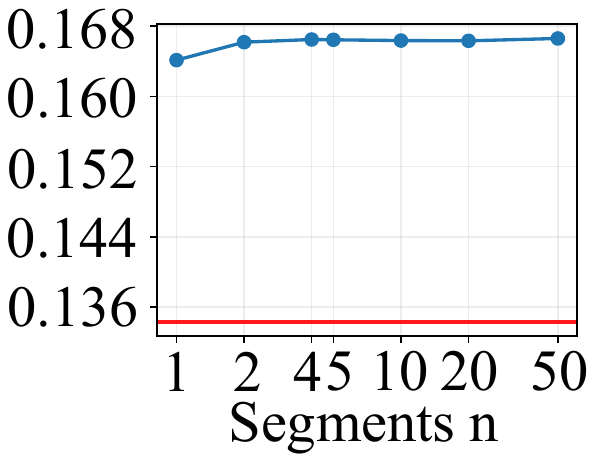}
        \vspace{-5mm}
        \caption{FGE$\downarrow$}
        \label{fig:fge_rh}
    \end{subfigure}
    \hfill
    \begin{subfigure}[b]{0.22\linewidth}
        \centering
        \includegraphics[width=\linewidth]{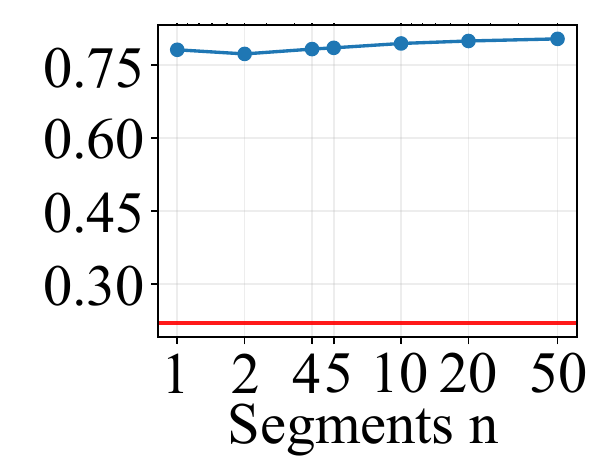}
        \vspace{-5mm}
        \caption{CR$\downarrow$}
        \label{fig:cr_rh}
    \end{subfigure}
    \hfill
    \begin{subfigure}[b]{0.22\linewidth}
        \centering
        \includegraphics[width=\linewidth]{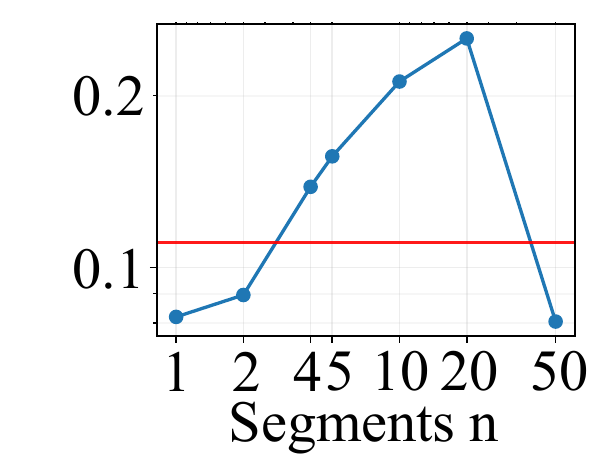}
        \vspace{-5mm}
        \caption{Curv (rad; log-scale y-axis)}
        \label{fig:curv_rh}
    \end{subfigure}
    \hfill
    \begin{subfigure}[b]{0.22\linewidth}
        \centering
        \includegraphics[width=\linewidth]{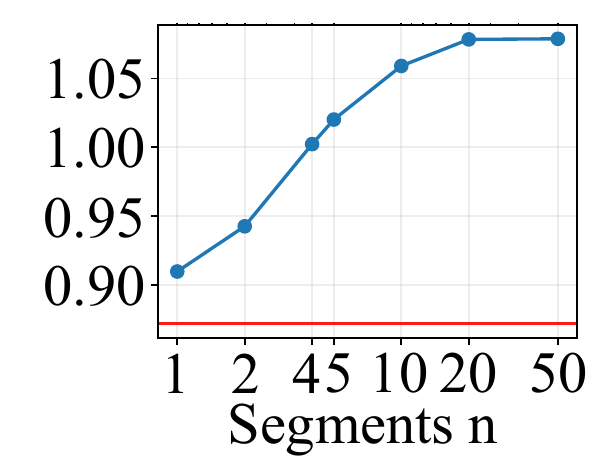}
        \vspace{-5mm}
        \caption{PLR}
        \label{fig:plr_rh}
    \end{subfigure}
  \vspace{-2mm}
  \caption{\textbf{Effect of pseudo receding-horizon segmentation $n$ on DP (DP-T-DDPM, 5 denoising steps, fixed seed 42).} The red horizontal line indicates CoFL with a fixed $100\times100$ query grid.}

  \label{fig:pseudo_rh_*}
  \vspace{-5mm}
\end{figure*}

\subsubsection{Problem Formulation and Trajectory Parameterization}
All DP-family baselines model a trajectory as a length-$T$ sequence of 2D displacements
$\Delta\mathbf{X}\in\mathbb{R}^{T\times 2}$ in normalized image coordinates, with $T=100$.
Waypoints are recovered by cumulative summation from the initial state:
\begin{equation}
\mathbf{X}=\mathbf{x}_0+\mathrm{cumsum}(\Delta\mathbf{X}).
\end{equation}
To improve optimization conditioning and keep the diffusion scale roughly unit-variance,
we rescale displacements by a dataset-level scalar computed once on the training set:
\begin{equation}
s_{\Delta}=\mathrm{median}\left(\left\|\Delta\mathbf{x}\right\|_2\right),
\end{equation}
where the median is taken over all per-step displacements $\Delta\mathbf{x}$ from all training trajectories
(i.e., over all entries of $\Delta\mathbf{X}$ across the entire training set). This single $s_\Delta$
is shared by all samples.

We train the diffusion model on normalized displacements
$\Delta\tilde{\mathbf{X}}=\Delta\mathbf{X}/s_{\Delta}$ and multiply back by $s_\Delta$ at inference:
\begin{equation}
\mathbf{X}=\mathbf{x}_0+\mathrm{cumsum}\!\left(s_{\Delta}\,\Delta\tilde{\mathbf{X}}\right).
\end{equation}

\subsubsection{Network Architecture}
\paragraph{Vision--language encoder (shared)}
All DP-family baselines reuse the same visual--language encoder as CoFL:
SigLIP2-B/16 at $224{\times}224$, followed by the same cross-modal fusion stack
(model dimension $d{=}768$, $8$ heads, $4$ fusion layers).
The encoder outputs a token sequence $\mathbf{C}\in\mathbb{R}^{N_v\times d}$ that conditions the denoiser.

\paragraph{Conditioning inputs (shared)}
Both denoisers are conditioned on the encoder tokens $\mathbf{C}$, the initial state $\mathbf{x}_0$,
and a diffusion time embedding, but differ in how the conditioning is injected
(pooled FiLM vs.\ token-level cross-attention).

\paragraph{Temporal U-Net denoiser (DP-C)}
We use the official 1D temporal U-Net denoiser with diffusion time embedding dimension 128
and three resolution stages with channel widths 256/512/1024.
The model uses kernel size 5 and group normalization with 8 groups.
Observations are provided as global conditioning:
the encoder tokens $\mathbf{C}$ are pooled (mean) into a single context vector, concatenated with $\mathbf{x}_0$
and the diffusion time embedding, then injected via FiLM modulation at multiple U-Net blocks.

\paragraph{Transformer denoiser (DP-T)}
We use the official transformer denoiser with 8 layers, 4 heads, and an embedding dimension $d{=}768$.
The initial state $\mathbf{x}_0$ and the diffusion time embedding are each projected to $d{=}768$
and appended as additional context tokens to $\mathbf{C}$, then injected via cross-attention
at multiple transformer blocks.

\subsubsection{Training Objective and Defaults}
\label{app:dp_training}

All DP variants share the same data preprocessing and displacement normalization, but use different training objectives depending on the sampler family.

\paragraph{DDPM objective (DP-*-DDPM)}
For stochastic reverse diffusion, we adopt the standard DDPM noise-prediction parameterization.
At a discrete noise level $n\in\{1,\ldots,N_{\mathrm{diff}}\}$ with variance schedule $\{\beta_n\}$
and $\bar{\alpha}_n=\prod_{s=1}^{n}(1-\beta_s)$, the forward process is
\begin{equation}
\Delta\tilde{\mathbf{X}}_n
= \sqrt{\bar{\alpha}_n}\,\Delta\tilde{\mathbf{X}}
+ \sqrt{1-\bar{\alpha}_n}\,\epsilon,\quad
\epsilon\sim\mathcal{N}(0,\mathbf{I}).
\end{equation}
The denoiser predicts the injected noise
$\hat{\epsilon}=\epsilon_\theta(\Delta\tilde{\mathbf{X}}_n,n\mid\mathbf{C},\mathbf{x}_0)$
and is trained with the classic diffusion policy loss
\begin{equation}
\mathcal{L}_{\mathrm{DDPM}}
=
\mathbb{E}_{\Delta\tilde{\mathbf{X}},\,n,\,\epsilon}
\left[
\left\|
\epsilon - \epsilon_\theta(\Delta\tilde{\mathbf{X}}_n, n \mid \mathbf{C}, \mathbf{x}_0)
\right\|_2^2
\right].
\end{equation}

\paragraph{Flow-matching objective (DP-*-Flow/ODE)}
For ODE-style sampling, we train a velocity field via flow matching with a continuous time
$t\in(0,1)$ (sampled per batch).
We draw $\mathbf{z}\sim\mathcal{N}(0,\mathbf{I})$ and construct a linear interpolation in displacement space:
\begin{equation}
\Delta\tilde{\mathbf{X}}_{t} = t\,\mathbf{z} + (1-t)\,\Delta\tilde{\mathbf{X}} ,
\end{equation}
with the constant target velocity (standard flow matching)
\begin{equation}
\mathbf{u}_{t} = \mathbf{z} - \Delta\tilde{\mathbf{X}} .
\end{equation}
The model predicts velocity
$\mathbf{v}_\theta(\Delta\tilde{\mathbf{X}}_{t}, t \mid \mathbf{C}, \mathbf{x}_0)$
and is trained by regression:
\begin{equation}
\mathcal{L}_{\mathrm{FM}}
=
\mathbb{E}_{\Delta\tilde{\mathbf{X}},\,t,\,\mathbf{z}}
\left[
\left\|
\mathbf{u}_{t} - \mathbf{v}_\theta(\Delta\tilde{\mathbf{X}}_{t}, t \mid \mathbf{C}, \mathbf{x}_0) 
\right\|_2^2
\right].
\end{equation}

All variants use batch size $32$ and AdamW (lr $10^{-4}$, weight decay $10^{-5}$) for $50$ training epochs,
with a cosine learning-rate schedule and $40000$ warmup steps.
For all reported results, we select the checkpoint with the lowest validation loss.

\subsubsection{Inference-Time Sampling}
\label{app:dp_inference}
At test time, we sample Gaussian noise in displacement space and map it to a displacement sequence
$\Delta\tilde{\mathbf{X}}$ using the sampler corresponding to each training objective, then recover waypoints
$\mathbf{X}$ by rescaling and cumulative summation.

\paragraph{DP-DDPM (stochastic reverse diffusion)}
Let $\Delta\tilde{\mathbf{X}} \in \mathbb{R}^{T\times 2}$ denote the
full displacement sequence of length $T$ (each step a $d$-DoF displacement),
and diffusion is performed on this entire sequence.
We initialize $\Delta\tilde{\mathbf{X}}_{N_{\mathrm{diff}}}\sim\mathcal{N}(0,\mathbf{I})$ and run the standard
discrete reverse process for $N_{\mathrm{diff}}$ steps.
At each noise level $n\in\{1,\dots,N_{\mathrm{diff}}\}$, the denoiser
$\epsilon_\theta(\Delta\tilde{\mathbf{X}}_n,n\mid\mathbf{C},\mathbf{x}_0)$
is used to compute the reverse transition and obtain $\Delta\tilde{\mathbf{X}}_{n-1}$,
until reaching $\Delta\tilde{\mathbf{X}}_0$.

\paragraph{DP-Flow/ODE (deterministic ODE integration)}
Similarly, Flow/ODE variants operate on the entire length-$T$ displacement sequence
$\Delta\tilde{\mathbf{X}}_t \in \mathbb{R}^{T\times 2}$.
The model predicts a velocity field
$\mathbf{v}_\theta(\Delta\tilde{\mathbf{X}}_{t}, t \mid \mathbf{C}, \mathbf{x}_0)$ trained by flow matching.
We initialize with pure noise
$\Delta\tilde{\mathbf{X}}_{1}\sim\mathcal{N}(0,\mathbf{I})$
and integrate the ODE from $t{=}1$ to $t{=}0$:
\begin{equation}
\frac{d\,\Delta\tilde{\mathbf{X}}_{t}}{dt}
=
\mathbf{v}_\theta(\Delta\tilde{\mathbf{X}}_{t}, t \mid \mathbf{C}, \mathbf{x}_0),
\end{equation}
using $N_\text{ode}$ Euler steps to obtain $\Delta\tilde{\mathbf{X}}_{0}$.
We then recover $\mathbf{X}$ by multiplying $s_\Delta$ and applying cumulative summation.

\section{More Examples of Trajectories on Validation Set}
\label{app:more_example_benchmark}
Additional qualitative examples corresponding to Fig.~\ref{fig:comparision} are provided in Fig.~\ref{fig:example_benchmark}.

\section{Pseudo Receding-Horizon Protocol for Diffusion-Policy Evaluation}
\label{app:pseudo_rh}

\subsection{Motivation}
A potential concern is that DP may be disadvantaged in our benchmark because we execute each predicted action chunk in an open-loop rollout, whereas diffusion-based controllers are often deployed with frequent closed-loop replanning.
If this evaluation-mode mismatch were the dominant factor, then increasing the replanning frequency---i.e., reconditioning DP on intermediate states more often---should yield a clear and systematic improvement in accuracy and safety, thereby narrowing the gap to CoFL.

To isolate this factor {without} introducing additional environment feedback, learned critics, or controller-specific heuristics, we design a fully offline {pseudo receding-horizon} (pseudo-RH) protocol.
Crucially, pseudo-RH does not alter the observation stream or add extra supervision; it only changes how often DP is re-queried from updated states during the offline rollout.
This turns replanning into a controlled variable for diagnosing whether DP's weakness is primarily an artifact of open-loop execution.

\subsection{Protocol}
Let DP output an $T$-step action chunk per query (we use $T{=}100$ throughout).
We split the rollout into $n$ segments.
For segment $i\in\{1,\dots,n{-}1\}$, we query DP conditioned on the current segment start state and execute only the first $h=\lfloor T/n \rfloor$ actions, discarding the remainder; the next segment then re-queries DP from the resulting state.
For the final segment, we execute the full $T$-step prediction rather than a truncated prefix, avoiding an artificial shortening of goal-reaching behavior and allowing DP to terminate at its own predicted endpoint.
We sweep $n$ and report FGE, CR, Curv, and PLR under this protocol (Fig.~\ref{fig:pseudo_rh_*}).

\subsection{Model/Configuration Details}
CoFL serves as a fixed reference (red horizontal line in Fig.~\ref{fig:pseudo_rh_*}) using a dense query grid of $100\times100$ and 100-step Euler rollout.
For DP, we use the DP-T-DDPM variant and sample with $5$ denoising steps for all $n$.
To reduce stochastic variation as a confounder, we fix the diffusion noise seed to $42$ for all runs reported in this appendix.
All other settings follow those used in the main experiments.

\subsection{Interpretation}
If DP were substantially hindered by open-loop execution, increasing $n$ would be expected to yield a marked and monotonic improvement: lower FGE and CR, reduced curvature, and higher path-length progress.
Instead, we observe that DP's metrics change only mildly with $n$, without a consistent trend indicating that replanning is the missing ingredient.
CoFL remains advantageous under its fixed inference configuration.
These results suggest that the primary performance gap in our main results is unlikely to be explained by evaluating DP in an overly open-loop regime.

Beyond serving as a diagnostic control, the pseudo-RH results further corroborate our main text analysis.
Diffusion-style trajectory generation remains an effective way to model continuous action sequences; however, in this task its supervision is still trajectory-local, constraining finite-horizon rollouts from observed starts rather than dense control behavior over the workspace.
The limited gains under pseudo-RH suggest that the key difference is not merely execution frequency, but whether the model can be supervised with workspace-level guidance.
In contrast, CoFL converts each scene--instruction pair into dense, locally grounded control supervision, enabling spatially consistent queries from arbitrary workspace locations.

\end{document}